# Turning hazardous volatile matter compounds into fuel by catalytic steam reforming: An evolutionary machine learning approach

Alireza Shafizadeh[1,2,†], Hossein Shahbeik[1,3,†], Mohammad Hossein Nadian[4], Vijai Kumar Gupta[5,6], Abdul-Sattar Nizami[7], Su Shiung Lam[3,1,8], Wanxi Peng[1,*], Junting Pan[9,*], Meisam Tabatabaei[3,1,10,*], Mortaza Aghbashlo[2,1,*]

[1] Henan Province Engineering Research Center for Forest Biomass Value-added Products, School of Forestry, Henan Agricultural University, Zhengzhou, 450002, China
[2] Department of Mechanical Engineering of Agricultural Machinery, Faculty of Agricultural Engineering and Technology, College of Agriculture and Natural Resources, University of Tehran, Karaj, Iran
[3] Higher Institution Centre of Excellence (HICoE), Institute of Tropical Aquaculture and Fisheries (AKUATROP), Universiti Malaysia Terengganu, 21030 Kuala Nerus, Terengganu, Malaysia
[4] School of Cognitive Sciences, Institute for Research in Fundamental Sciences (IPM), Tehran, Iran
[5] Biorefining and Advanced Materials Research Center, Scotland's Rural College (SRUC), Kings Buildings, West Mains Road, Edinburgh, EH9 3JG, UK
[6] Centre for Safe and Improved Food, Scotland's Rural College (SRUC), Kings Buildings, West Mains Road, Edinburgh, EH9 3JG, UK
[7] Sustainable Development Study Centre (SDSC), Government College University, Lahore, Pakistan
[8] University Centre for Research and Development, Department of Chemistry, Chandigarh University, Gharuan, Mohali, Punjab, India
[9] State Key Laboratory of Efficient Utilization of Arid and Semi-arid Arable Land in Northern China, Institute of Agricultural Resources and Regional Planning, Chinese Academy of Agricultural Sciences, Beijing 100081, China
[10] Department of Biomaterials, Saveetha Dental College, Saveetha Institute of Medical and Technical Sciences, Chennai 600 077, India

*Correspondence:

Mortaza Aghbashlo (maghbashlo@ut.ac.ir)
Meisam Tabatabaei (meisam_tab@yahoo.com)
Junting Pan (panjunting@caas.cn)
Wanxi Peng (pengwanxi@henau.edu.cn)

†These authors contributed equally.

**Abstract**

Chemical and biomass processing systems release volatile matter compounds into the environment daily. Catalytic reforming can convert these compounds into valuable fuels, but developing stable and efficient catalysts is challenging. Machine learning can handle complex relationships in big data and optimize reaction conditions, making it an effective solution for addressing the mentioned issues. This study is the first to develop a machine-learning-based research framework for modeling, understanding, and optimizing the catalytic steam reforming of volatile matter compounds. Toluene catalytic steam reforming is used as a case study to show how chemical/textural analyses (e.g., X-ray diffraction analysis) can be used to obtain input features for machine learning models. Literature is used to compile a database covering a variety of catalyst characteristics and reaction conditions. The process is thoroughly analyzed, mechanistically discussed, modeled by six machine learning models, and optimized using the particle swarm optimization algorithm. Ensemble machine learning provides the best prediction performance ($R^2 > 0.976$) for toluene conversion and product distribution. The optimal tar conversion (higher than 77.2%) is obtained at temperatures between 637.44 and 725.62 °C, with a steam-to-carbon molar ratio of 5.81–7.15 and a catalyst BET surface area of 476.03–638.55 $m^2$/g. The feature importance analysis satisfactorily reveals the effects of input descriptors on model prediction. Operating conditions (50.9%) and catalyst properties (49.1%) are equally important in modeling. The developed framework can expedite the search for optimal catalyst characteristics and reaction conditions, not only for catalytic chemical processing but also for related research areas.

**Nomenclatures:**

| | |
|---|---|
| $A$ | Surface area under the peak. |
| $CI$ | Crystallinity index |
| D | Average crystal size |
| $k$ | Scherrer constant |
| nm | Nanometer |
| x | Coordinate of the Gaussian curve |
| $y$ | Coordinate of the Gaussian curve |
| $y_0$ | Height of the peak starting point |
| $x_c$ | Peak position |
| $\beta$ | Full width at half maximum radians |
| $\theta$ | Braggs angle or peak position |
| $w$ | Constant number |
| $R^2$ | Coefficient of determination |
| $r$ | Correlation coefficient |

**Abbreviations**

| | |
|---|---|
| ANFIS | Adaptive neuro-fuzzy inference system |
| ANN | Artificial neural network |
| BET | Brunauer-Emmett-Teller |
| EML | Ensemble machine learning |
| GAM | Generalized additive model |
| GHSV | Gas hourly space velocity |
| GPR | Gaussian process regression |
| MOPSO | Multi-objective particle swarm optimization |
| MAE | Mean absolute error |
| RMSE | Root-mean-square error |
| PCA | Principal component analysis |
| PCs | Principal components |
| PSO | Particle swarm optimization |
| SD | Standard deviation |
| SHAP | SHapley Additive exPlanations |
| SVM | Support vector machine |
| WHSV | Weight hourly space velocity |
| XRD | X-ray diffraction |

# 1. Introduction

Energy production from biomass has attracted global attention because of growing concerns about the adverse impacts of fossil fuel utilization on the environment and human health. The annual primary biomass production is about 220 billion tonnes dry matter, amounting to 4,500 EJ energy (almost 10 times the world's primary energy consumption) (El Bassam, 2010). The potential of biomass resources to mitigate greenhouse gas emissions and achieve net-zero carbon dioxide, as the amount absorbed during the growing period counterbalances the emissions during biomass burning, makes them an attractive option for energy and chemical production. Moreover, the indirect utilization of biomass energy through converting biomass into biofuel demonstrates the potential of biomass to play a significant role in the world's energy future (Long et al., 2013).

A variety of biochemical and thermochemical pathways can be applied to convert biomass into biofuels and biochemicals (Soltanian et al., 2020). Generally, biochemical routes are more fitted for biomass rich in starch and sugar (Parthasarathy et al., 2017). Due to the inherent resistance of cell walls to microbial and enzymatic breakdown, lignocellulosic biomass feedstocks have a recalcitrant structure that makes their biochemical conversion expensive, chemically challenging, and time-consuming (Huang et al., 2018). Unlike biochemical pathways, thermochemical routes are quick and can handle a broad spectrum of biomass feedstocks without requiring chemical pretreatment.

Thermochemical conversion techniques include combustion, torrefaction, pyrolysis, gasification, and hydrothermal processing (Seo et al., 2022). Pyrolysis and gasification are advantageous thermochemical technologies for producing biofuels and biochemicals in an economically feasible and energetically efficient manner (Azizi et al., 2020). However, tar formation is a challenging issue in implementing these processes (particularly gasification) on a large scale (Li et al., 2015). Tar formation blocks pipelines, damages reactors, decreases process efficiency, causes intensive energy loss, and increases operating costs (Anis et al., 2013; Rahman et al., 2020). Besides technical problems, tar can cause serious environmental and health problems if released into the atmosphere (Yamaguchi et al., 2002). Hence, tar formation should be avoided, and/or the generated tar should be treated using effective techniques for technical, economic, environmental, and safety reasons.

Tar is generally generated in the pyrolysis step of thermochemical processes when lignocellulosic substances are decomposed because of thermal treatment. The quantity and composition of tar during thermochemical processes can be influenced by various factors, such as biomass composition, reaction temperature, reaction agent, reaction pressure, and catalyst type/loading (Ren et al., 2022). Tar is a condensed mixture of organic molecules, hydrocarbons (single-ring to five-ring aromatic and polycyclic compounds), and oxygenates compounds (Cortazar et al., 2018). Tar is defined in three ways: 1. The extracted organic substance (mostly largely aromatics) from organic materials under thermal or partial oxidation conditions, 2. The compound of condensable hydrocarbons contains monocyclic to bicyclic and polycyclic aromatics plus other oxygenated groups, 3. Hydrocarbons with molecular weight higher than benzene (J. Ren et al., 2020).

Upon temperature, three classes of tar are usually generated during biomass decomposition. The first class, including levoglucosan, furfural, and hydroxyacetaldehyde, is produced at 400–700 °C. The second class, including phenolic and olefin compounds (phenol, cresol, and xylene), is generated at 700–850 °C. The third class, including complex aromatic compounds (i.e., benzene, naphthalene, pyrene, and toluene), is produced at around 850–1000 °C. Typically, one-ring (e.g., toluene) and two-ring (e.g., naphthalene) aromatic hydrocarbons account for approximately 75% of tar content. Toluene is the highest abundant compound in tar composition (Guan et al., 2016). Accordingly, toluene has been considered as a model compound in many tar-related studies.

Besides being a tar model compound, a considerable amount of toluene is released from industries and vehicles into the environment every day. This highly toxic volatile organic compound can seriously threaten the ecological environment and human health (Hao et al., 2022). Reportedly, toluene is related to asthma, nasopharyngeal cancer, and other human health problems (Mo et al., 2009). Toluene is also one of the main compounds resulting in chemical smog (Wu et al., 2022). Accordingly, chemical and fuel industries urgently need to mitigate toluene emission and convert it into useful products (de Lasa et al., 2011; Mo et al., 2021).

Several physical and thermochemical approaches have been introduced to deal with tar. The physical approaches include wet scrubbing, cyclone separation, electrostatic precipitation, and heat exchanging. The thermochemical methods are catalytic steam reforming, thermal cracking, plasma reforming, and partial oxidation (Xu et al., 2022). Catalytic steam reforming has gained more attention because of its strong capability to convert tar into hydrogen-rich syngas (Ortiz-Toral et al., 2011). Several reaction pathways occurring during catalytic tar steam reforming are shown in Figure 1.

Generally, catalytic steam reforming reactions can be classified into primary and secondary reactions, predominantly related to heterogeneous catalytic and gas-phase reactions, respectively. The primary reactions are highly endothermic and are thus more favorable at higher temperatures. However, the exothermic nature of the water-gas shift reaction ($CO + H_2O \rightleftharpoons CO_2 + H_2$) can hinder the formation of hydrogen at elevated temperatures. The secondary reactions include the thermal cracking of both hydrocarbons and oxygenated compounds and various interconversion reactions.

These reactions are more likely to occur under intense operating conditions, such as prolonged residence times and high temperatures (Santamaria et al., 2021).

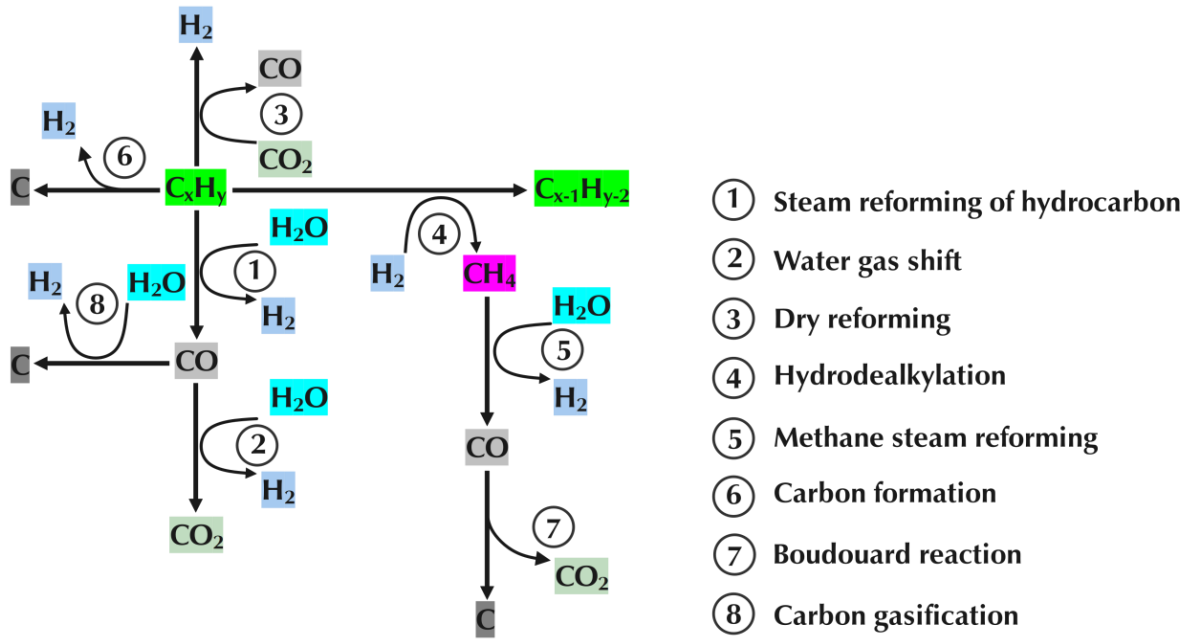

Figure 1. Possible reaction pathways during tar steam reforming (Tan et al., 2020b). Redrawn with permission from Springer Nature. Copyright© 2020.

The choice of the reactor can have a significant impact on the effectiveness of the catalytic steam reforming process (Lopez et al., 2022). Various reactor configurations have been introduced for in-situ or ex-situ steam reforming of tar, including two-stage fixed bed systems (Cao et al., 2014; Chen et al., 2016; Dharma et al., 2016; Gao et al., 2021b) and fluidized bed-fixed bed systems (Xiao et al., 2011). These configurations can be coupled with other reactor types, such as continuous screw kiln reactors (Efika et al., 2012), spouted bed-fluidized bed reactors (Arregi et al., 2018; Santamaria et al., 2020), and conical spouted bed reactors (Fernandez et al., 2022), to reform the volatile matter produced during biomass pyrolysis or gasification. However, the process faces technical challenges such as catalyst deactivation (Liu et al., 2017) and low mechanical strength and instability of the catalyst within reactors (e.g., fixed bed reactors) (Guan et al., 2016).

Applying highly active catalysts in steam reforming can significantly increase tar elimination while yielding more useful products (L. Ren et al., 2020). Catalysts can also lower the temperature required for effective reforming (Buentello-Montoya et al., 2019). Nevertheless, manufacturing efficient catalysts for the catalytic steam reforming of tar is costly and lengthy. After developing catalysts, various experimental measurements should be undertaken to understand and optimize the reforming process. Several catalyst- and operational-related parameters can affect the effectiveness of tar steam reforming. The catalyst-related parameters include catalyst type, loading, activity, crystallinity, crystal form, crystal size, specific surface area, and pore dimension (Wu et al., 2020). Influential operating parameters are total gas hourly space velocity (GHSV) or weight hourly space velocity (WHSV), steam-to-carbon molar ratio, carrier gas initial temperature, operating temperature, and processing time (Łamacz et al., 2019; Meng et al., 2017).

Understanding the complex relationships between influential parameters and tar conversion rate during the catalytic steam reforming process using the conventional trial-and-error methods is too cost-intensive and time-consuming. Innovative modeling methods can help better understand complex processes such as tar catalytic steam reforming without needing various experimental trials. Several modeling methods, including mathematical models (dos Anjos et al., 2020; Li et al., 2018), computational fluid dynamics (Marias et al., 2016), and kinetic and thermodynamic equilibrium models (Ahmed et al., 2015; Oemar et al., 2014; Vivanpatarakij and Assabumrungrat, 2013), have been used for modeling tar catalytic steam reforming. These modeling methods suffer from some weaknesses. For instance, computational fluid dynamics models are computationally intensive, requiring numerous assumptions and oversimplifications. The reaction mechanisms are sometimes unknown or not fully understood in kinetic models.

Kinetic models are also extremely reliant on the estimation of complex reaction rates. Thermodynamic modeling frequently assumes an equilibrium state and needs particular expensive software such as ASPEN Plus (Umenweke et al., 2022). These approaches do not always provide reliable and accurate results.

Advanced modeling techniques, including machine learning, have become increasingly popular for modeling complex phenomena that are difficult to program using traditional programming methods. Machine learning can think and learn based on input data sets without being explicitly programmed (Dhall et al., 2020). This modeling tool uses algorithms and techniques for automating solutions to complex and nonlinear systems with multivariant and uncertain phenomena (Aghbashlo et al., 2021). Machine learning can provide a chance to get hidden trends and patterns from past historical data. It can substantially reduce the efforts required to find the optimal catalyst configurations and operating conditions in tar catalytic steam reforming.

Despite the interesting features of machine learning technology, there is almost no study on using this powerful tool to characterize tar catalytic steam reforming. Therefore, this study is set to introduce a universal machine learning model to predict and optimize tar catalytic steam reforming. The present research aims to pave the way for machine learning technology in tar catalytic steam reforming. A hazardous tar model compound or a volatile organic compound (toluene) is considered in this study. The required experimental database is compiled from the published literature. Several statistical analyses and mechanistic explanations are applied to evaluate the relationship between the independent input features and the dependent desired responses. Machine learning is designed to predict toluene conversion and evolved syngas composition ($H_2$, $CO$, $CO_2$, $CH_4$) during tar catalytic steam reforming based on catalyst features and operating conditions. Six machine learning models, including artificial neural network (ANN), adaptive neuro-fuzzy inference system (ANFIS), generalized additive model (GAM), support vector machine (SVM), Gaussian process regression (GPR), and ensemble machine learning (EML), are used to model the process. Once the best model is chosen, optimal operating conditions and catalyst properties are sought using an evolutionary algorithm. Finally, feature importance analysis is applied to understand the effects of the independent input parameters on the prediction of dependent desired responses.

## 2. Research Methodology

Figure 2 outlines the steps taken in this study to model and optimize toluene catalytic steam reforming. First, a comprehensive review of published research articles was conducted, and eligible papers were selected for further screening. Second, process variables and catalyst features were extracted from the qualified papers. Third, the extracted data was used to train six machine learning models, including ANN, ANFIS, GAM, SVM, GPR, and EML. The particle swarm optimization (PSO) algorithm was then utilized to adjust the hyperparameters of the trained models. Fourth, the multi-objective particle swarm optimization (MOPSO) algorithm was applied to identify optimal operating conditions and catalyst characteristics. Fifth, feature importance analysis was performed using the best-performing machine learning model. Finally, the selected model was presented as a software platform to facilitate its use in future research and real-world applications.

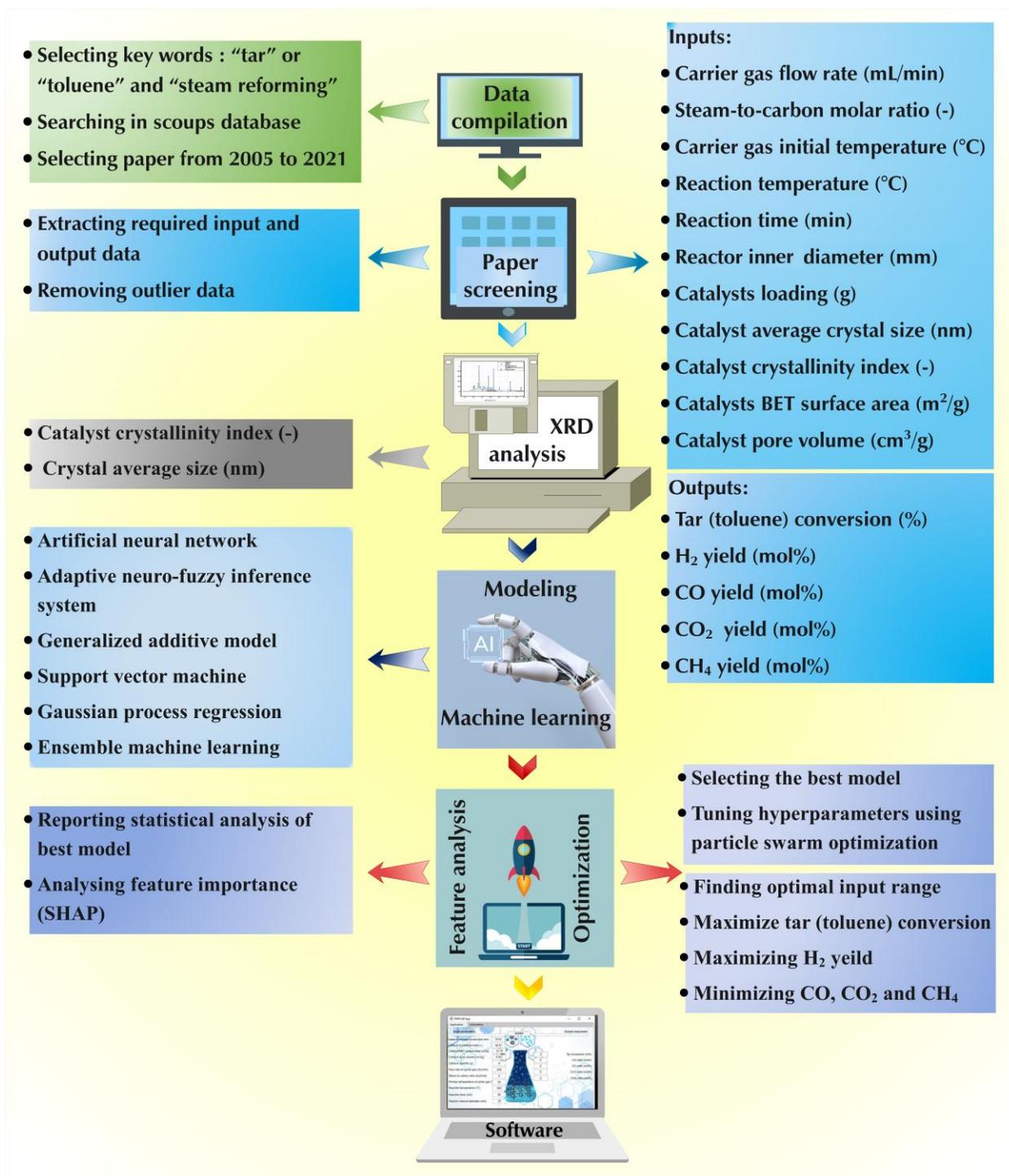

Figure 2. Research method used in the present study.

## 2.1. Literature survey and data compilation

The Scopus database was searched using the keywords "tar" or "toluene" and "steam reforming" in the article title, abstract, and keywords to compile the data. A total of 745 articles were initially retrieved, but only 349 papers published between 2005 and 2021 that met the research criteria were selected for further evaluation. Input descriptors and output variables were carefully selected based on the reported data, which is crucial for developing an accurate model. The chosen descriptors should directly impact the dependent responses and be independent of each other. When selecting input descriptors for a machine learning model, it is crucial to consider the availability of sufficient quantitative data, the coverage of a diverse range of variations, and the appropriate step size for the range of variation based on the nature of the input descriptors. A total of 820 data patterns were extracted from 25 articles (Table S1), and outliers were removed before modeling. A total of 592 data patterns were retained for data mining.

The independent input data extracted for the study were categorized into two general groups: catalyst textural characteristics and operating parameters. The catalyst textural characteristics group included catalyst BET surface area ($m^2/g$), pore volume ($cm^3/g$), crystallinity index (%), and average crystal size (nm). It is worth noting that catalyst chemical properties, such as the number of catalyst acid sites, were excluded from the modeling database due to a lack of quantitative data, mainly reported through temperature-programmed desorption graphs and qualitative discussions.

The operating parameters comprised steam-to-carbon molar ratio (-), reaction temperature (°C), carrier gas initial temperature (°C), processing time (min), catalyst loading (g), carrier gas flow rate (mL/min), and reactor inner diameter (mm). The steam-to-carbon molar ratio (-) and temperature (°C) are critical parameters that significantly influence the tar steam reforming process (Artetxe et al., 2017, 2016). Notably, an appropriate steam content during reforming can benefit tar conversion, while excessively high temperatures can result in bed clogging and rapid catalyst deactivation (Yue et al., 2010). Preheating tar at the reactor inlet can provide two advantages. Firstly, it reduces the time needed for reactants to reach the desired reaction temperature. Secondly, it minimizes the temperature difference between the tar and the reactor, which improves tar reforming (Chen et al., 2020). The initial temperature (°C) of the carrier gas in the model determines whether preheating is conducted during the tar reforming process or not. In addition, catalyst deactivation can occur during tar reforming, leading to a decline in tar conversion. The reaction time (in minutes) can effectively indicate the catalytic deactivation as the reaction progresses (Gao et al., 2021b).

GHSV is a critical and influential independent input parameter in tar catalytic steam reforming. GHSV is defined as the gas flow rate at the catalytic reactor outlet divided by the bulk volume of the catalyst bed (Rapagna, 1998). This factor indicates contact frequencies among the catalyst and the reactants or, in other words, equivalent to the residence time of reactants in contact with the catalyst (Claude et al., 2019). GHSV can influence tar steam reforming and change the composition of the evolved syngas (i.e., $H_2$ and $CO_2$) products (Chianese et al., 2016). However, GHSV has rarely been reported in tar catalytic steam reforming studies. Therefore, carrier gas flow rate and reactor inner diameter catalyst loading (g) were chosen as GHSV representatives to address this issue. The carrier gas refers to an inert gas, such as nitrogen or helium, used to transport the reactants through the reactor in catalytic reactions. The carrier gas flow rate can impact the reaction rate, selectivity, and other process parameters. GHSV varies by changing the total feedstock flow rate (Xu et al., 2015). In addition, the gas flow rate can be adjusted by altering the inner diameter of the reactor and the quantity of catalyst added to the reactor cylinder. Therefore, these variables can serve as alternatives to the need for GHSV as independent input parameters.

The dependent output parameters include tar conversion rate (%) or simply the amount of tar percentage converted, $H_2$ production (mol%), CO production (mol%), $CO_2$ production (mol%), and $CH_4$ production (mol%).

All input and output parameter data, except for crystallinity index (%) and average crystal size (nm), were extracted from reported numeric data in tables or plotted graphs using Getdata software for digitizing graphs. Notably, the crystallinity index and average size are important because of their significant effects on the performance and durability of a catalyst. A higher crystallinity index generally means a more ordered crystal structure, improving catalytic activity and selectivity (Tanabe and Cohen, 2010). Indeed, a well-ordered crystal structure can provide more active sites for reactions and help ensure that the reaction proceeds in the desired direction. In addition, larger particles can have a lower surface area, meaning fewer active reaction sites. Smaller particles, on the other hand, can have a higher surface area, leading to more active sites and potentially higher catalytic activity. However, very small particles may be more prone to agglomeration or deactivation over time.

The crystallinity index (%) and average crystal size (nm) of catalysts were obtained from X-ray diffraction (XRD) crystallography curves reported in the selected papers. High-resolution XRD graphs for the reported catalysts were downloaded and cleaned from redundant lines and elements to accomplish this. The cleaned XRD curves were digitized using a computer program (Digitizelt) to obtain their original (x,y) coordinates. This software can automatically extract the required number of unique coordinates. The produced coordinates were exported into an Excel file. Next, the Excel file data was imported into Origin software, and the XRD graphs were plotted in the software (Figure 3 (A)). In the next step, each individual plotted XRD graph was analyzed through the "Peak analyzer" tab in Origin software to extract the average crystal size and crystallinity index of the catalyst (Figure 3 (B)). The average crystal size (D, nm) and crystallinity index (CI, %) were computed based on Eqs. 1 and 2, respectively (Muniz et al., 2016).

$$D = \frac{k*\gamma}{\beta*\cos\theta} \tag{1}$$

$$CI = \frac{Area\ of\ all\ crysaliity\ peaks}{Area\ of\ all\ peaks\ (crystal + amorphous)} * 100 \tag{2}$$

where $k$ is the Scherrer constant (0.15406) for spherical crystallite with cubic symmetry ((wavelength of the X-ray source), $\beta$ is the full width at half maximum radians, and $\theta$ is Braggs angle or peak position (radians). It must be noted that Eq. 1 can be used for crystal sizes smaller than 200 nm.

The Gaussian fitting function was employed to calculate the average crystal size based on the Scherrer equation (Eq. 3). In this calculation, full width at half maximum was computed using the Gaussian fitting function by Origin software.

$$y = y_0 + \frac{A}{w\sqrt{\frac{\pi}{2}}} e^{-2\frac{(x-x_c)^2}{w^2}} \tag{3}$$

where $x$ and $y$ stand for the coordinates of the Gaussian curve, respectively. $y_0$ denotes the height of the peak starting point, $x_c$ shows the peak position, $w$ is a constant equal to 0.85, and $A$ indicates the surface area under the peak. The extracted data (823 data patterns) can be found in the "Supplementary Excel File".

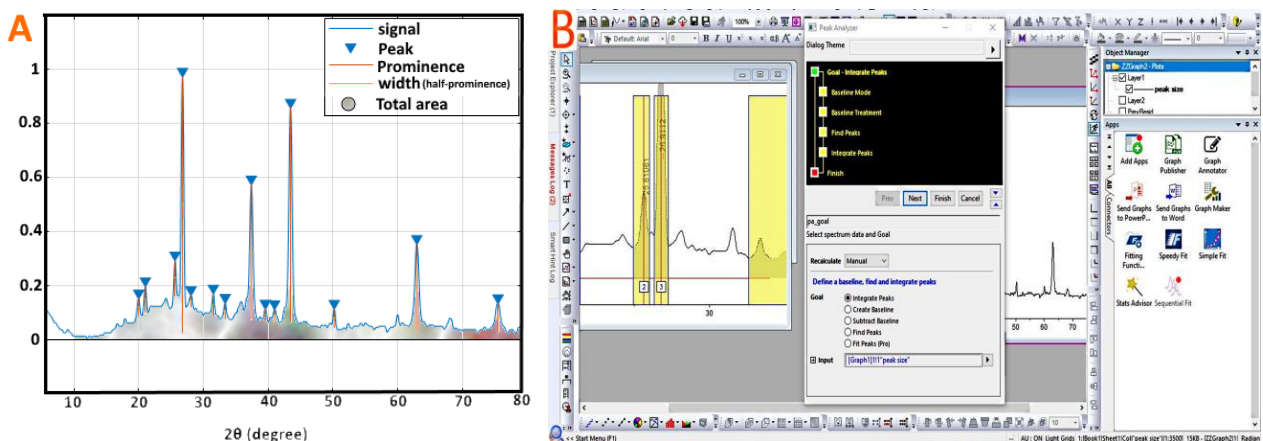

Figure 3. Schematic illustration of digitizing an XRD graph, manually selecting peaks to calculate crystal size and crystalline index of the catalysts (A), and analyzing peaks in Origin software (B).

## 2.2. Machine learning models

### 2.2.1 Modeling and optimizing tar catalytic steam reforming

This study applied six machine learning models (ANN, ANFIS, GAM, SVM, GPR, and EML) to predict the toluene conversion rate and the composition of resultant syngas ($H_2$, CO, $CO_2$, and $CH_4$) as a function of reaction conditions and catalyst characteristics. ANN is a highly parallel computing system comprising fully connected artificial neurons to mimic the learning and generalization ability of the human brain. ANN structure includes an input neuron layer, at least one hidden layer, and an output layer. During training, a numeric value is assigned for each connection (weight) and each neuron (bias) (Sarkar et al., 2021). The weights and biases of an ANN model can be adjusted using various activation functions (i.e., tangent, sigmoid, and linear). This model can provide solutions for highly complex systems thanks to its capability to solve problems in a stochastic manner.

ANFIS is an adaptive network type that integrates the principles of neural networks and fuzzy logic. This machine learning model has five fully connected layers: fuzzification, rules, normalization, defuzzification, and summation (Üstün et al., 2020). ANFIS has both positive aspects of the fuzzy logic approach (interpretability) and the supervised learning algorithm (training capability). It can also eliminate some drawbacks of lonely-used neural networks and fuzzy logic approaches. ANFIS operates like a feed-forward backpropagation network (Aghbashlo et al., 2019). This model can process complex and nonlinear information while having the capability to work in a parallel, distributed, and adaptable way.

GAM is a statistical framework similar to regression models. This approach can handle nonlinear and non-monotonic relationships that cannot be easily defined using nonlinear models. The interpretable and flexible hybrid GAM model combines generalized linear and additive models to include nonparametric smoothing terms. In better words, GAM uses a linear combination of multiple smooth functions of explanatory variables to model the response of interest (Shafizadeh et al., 2022). Accordingly, this model not only can provide predicted values but also can help understand the effect of the co-variables on the response variables (Shafizadeh et al., 2022). Generally, GAM can model complex problems without special conditions or unique solutions.

SVM is a powerful machine learning approach that utilizes a unique statistical basis to handle linear and nonlinear modeling problems in various fields. The core idea of SVM is to construct a hyperplane or a collection of hyperplanes in a high-dimensional space using an appropriate kernel function, which enables the separation of data into groups or the formation of a regression line. SVM can effectively handle complex datasets, even those with many or infinite dimensions, due to its utilization of kernel functions. The optimal line or hyperplane is selected by maximizing the margin between the closest training point and the hyperplane. SVM is a suitable model for classification and regression problems because of its ability to handle small data samples, high adaptability, and good generalization performance (Raghavendra. N and Deka, 2014).

GPR is a widely used supervised statistical machine learning algorithm that optimizes an unknown function over a training set (Lüder and Manzhos, 2020). This model provides a confidence interval for each point in the prediction, which quantifies the uncertainty of the forecast. GPR's structure is based on a nonparametric probabilistic kernel model that considers a distribution of input vectors and computes their probability. Instead of a scalar mean and variance, the computation involves a mean and covariance vector (Sharifzadeh et al., 2019). GPR is a flexible and parameter-free method that only requires the adjustment of hyperparameters. Additionally, GPR does not restrict the search space in any optimization scheme (Lüder and Manzhos, 2020). As a result, this model can be applied to various regression problems and predictions.

EML uses multiple machine learning algorithms and selects the best predictive results with various voting mechanisms (instead of using a fixed algorithm). This model exploits an aggregation function (aggregates K inducers) from *n* examples and *m* features to predict a single output. EML can provide excellent predictive performance for several reasons, including overfitting avoidance, local minim elimination, better fitting to data space, avoiding class imbalance, and solving the curse of dimensionality (Sagi and Rokach, 2018). In this model, the bias trends are decreased dramatically before remaining steady, while the variance trends are held steady before rising heavily (Dong et al., 2020). The learning error is continuously decreased until reaching the lowest level, followed by a rapid upward trend when the model becomes more complex (Dong et al., 2020).

PSO is a population-based stochastic optimization algorithm inspired by the intelligent social behavior of some animals, such as flocks of birds or schools of fish. This algorithm finds optimal points in a swarm by updating individual particles (velocity and position) from iteration to iteration. In the swarm, each individual is considered a particle with the potential solution of the optimized problem in D-dimensional solution space. The particle can memorize the optimal local and global position and velocity of the swarm. In every generation, the velocity of each dimension is simultaneously adjusted based on the combination of previous particle information and the new position of the particle. Until access to the optimal state, the states of the particles are constantly changed in the multi-dimensional search space of the algorithm. PSO has several advantages, including performing simple space and time computations, sensing the quality change in the environment and the response, diverse response capability, great stability, and intelligence adaptability (Wang et al., 2018). PSO can provide better solutions and faster convergence for irregular, noisy, or dynamically altered problems than the other heuristic approaches (Shafizadeh et al., 2022).

**2.2.2. Model development and optimization**

Figure 4 indicates the flowchart used to develop machine learning approaches for toluene tar catalytic steam reforming. The extracted data were normalized between zero and one and then introduced into machine learning models. It is vital in machine learning to assess the performance of the models unbiasedly to approve their generalization capability. Therefore, the k-fold cross-validation method was applied (Jung and Hu, 2015). This robust data resampling approach can also eliminate the possibility of overfitting in developed machine learning models (Suvarna et al., 2022). The compiled data patterns are equally split into k parts and used alternately in the training and testing stages. The models considered are trained using k-1 folds and tested using the last remaining one. This procedure is iterated k time to ensure that all the compiled data patterns are used in models' training and testing (de Rooij and Weeda, 2020). In addition, the k-fold cross-validation method can provide a reasonable estimation error in each learning algorithm (Li et al., 2020). This study used five-fold (k=5) cross-validation to train the considered machine learning models. The models are trained using approximately 80% of the randomly shuffled data patterns and tested using the remaining data points (≈20%).

Optimizing the hyperparameters of machine learning models is another effective strategy for improving their fitting performance. The PSO algorithm was selected for hyperparameter optimization in the present study. The parameters of the PSO algorithm used in the optimization process are presented in Table S2 (Supplementary Word File). Four statistical parameters, including coefficient of determination ($R^2$), mean absolute error (MAE), and root-mean-square error (RMSE), were used to evaluate the accuracy of the developed models. The best model was chosen by maximizing the $R^2$ value and minimizing the MAE and RMSE values. Once the best model was chosen, its outputs (objective functions) were fed into the MOPSO algorithm to optimize the process. The optimization was conducted to determine the optimal catalyst characteristics and operating conditions to achieve maximum tar conversion and $H_2$ yield and minimum CO, $CO_2$, and $CH_4$ yield. The parameters of the MOPSO algorithm used in the optimization process are presented in Table S3 (Supplementary Word File).

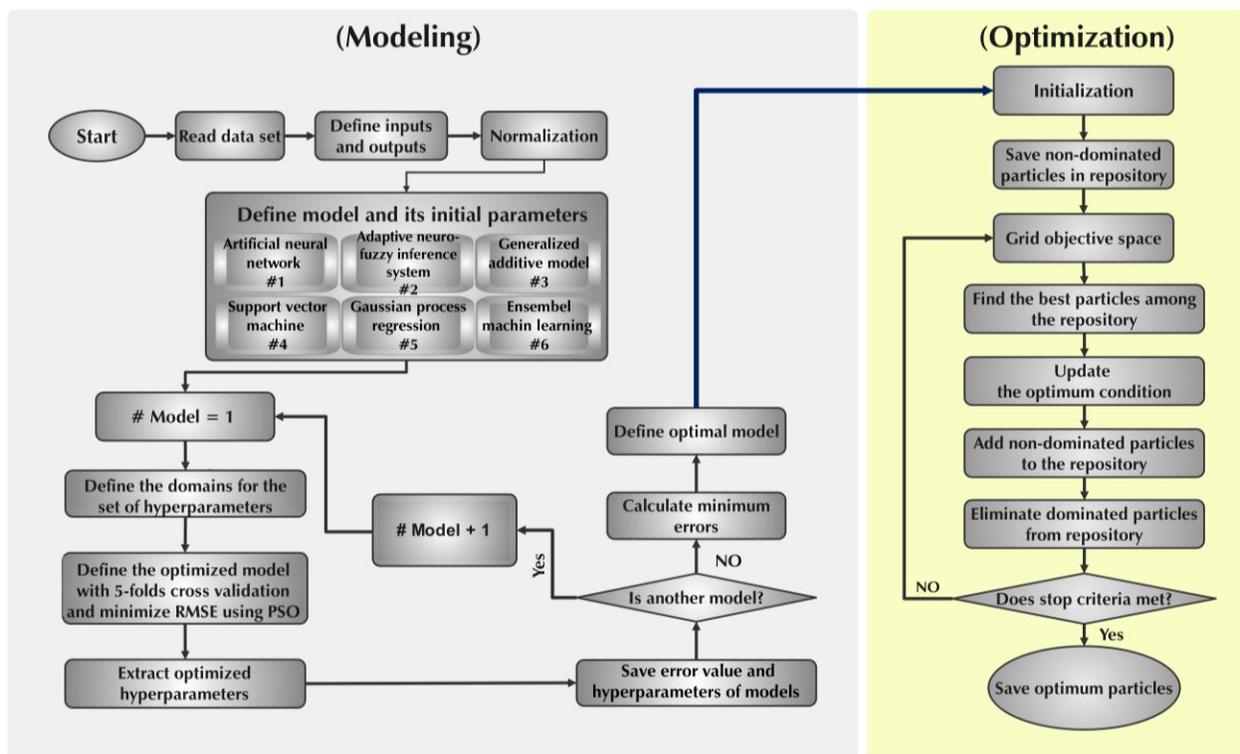

Figure 4. Flowchart used to develop machine learning approaches for modeling tar catalytic steam reforming.

## 2.3. Feature importance analysis

SHapley Additive exPlanations (SHAP) analysis was used to understand the importance of independent input parameters, including operating conditions and catalyst characteristics, in predicting the tar conversion process and the distribution of output products. This feature importance analysis method is based on game theory, which computes the impact of every subset of a given feature dataset on a model prediction. SHAP analysis approximates the feature importance values through special weighted linear regression (Marcilio and Eler, 2020). This analysis orders the features based on the mean absolute or the maximum absolute value of the SHAP values for each feature. In this study, the features were ordered based on the maximum absolute value because of the capability of this method to find features with high impacts on individual parameters. In addition, the overall importance of operating conditions and catalyst characteristics were assessed on each corresponding output.

## 3. Results and discussion

### 3.1. Data analysis

A descriptive analysis of all the input and output variables is conducted to acquire preliminary insights into tar catalytic steam reforming. Figure 5 depicts the box plot analysis results for the input/output variables. The data description is given on the left side, including box range, standard deviation, and mean value, while specific data points and kernel distribution are shown on the right side. The catalyst properties, including average crystal size, crystallinity index, BET surface area, and pore volume, are distributed between 3.33–25.3 nm, 4.44–94.03%, 5.7–322 $m^2/g$, and 0.02–0.91 $cm^3/g$, respectively. This wide range of catalyst properties could be attributed to the tremendous diversity of catalysts (i.e., nickel-, alkali-, acid-based, and natural mineral catalysts) employed in tar catalytic steam reforming.

It should be noted that selecting proper catalysts with high activity and good resistivity against coke deposition is the key factor in the reliability and effectiveness of reforming processes (Gao et al., 2022). In the compiled data, reaction temperature and steam-to-carbon molar ratio vary from 300 to 900 °C and 0.5 to 8, respectively. As a result, the compiled data can encompass a wide range of feasible operating conditions for the tar catalytic steam reforming process. The mean values for carrier gas flow rate and reaction are 16.7 mL/min and 346 min, respectively.

According to the database, tar conversion ranges between 0.82 and 99.85%. The large variations in the tar conversion percentage could be attributed to chemically and structurally diverse catalyst materials and various reaction conditions reported in the published literature. $H_2$

has the highest contribution to resultant syngas (0.70–98.8 mol%), followed by CO (0.10–63.3 mol%), $CO_2$ (0.36–28.2 mol%), and $CH_4$ (0.0–28.0 mol%). Generally, the input and output variables are distributed within reasonable ranges without being overly concentrated or widely separated. Such an inclusive database could make the generalization and optimization of tar catalytic steam reforming possible using machine learning technology.

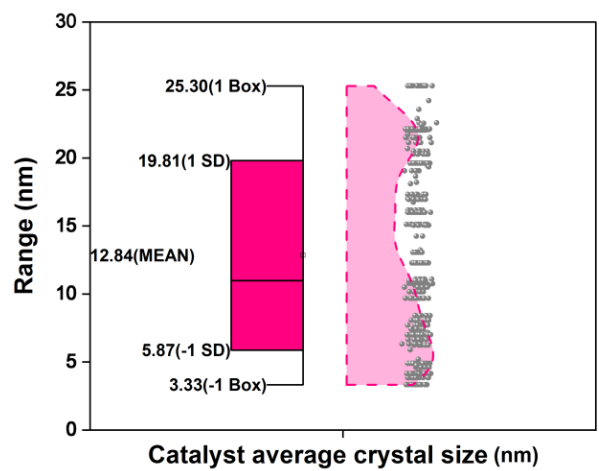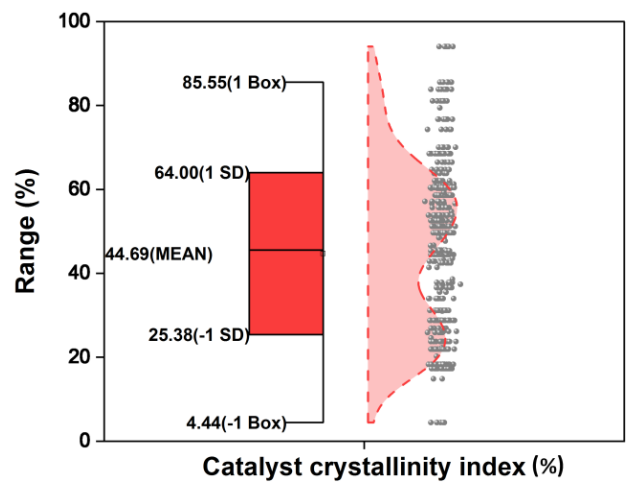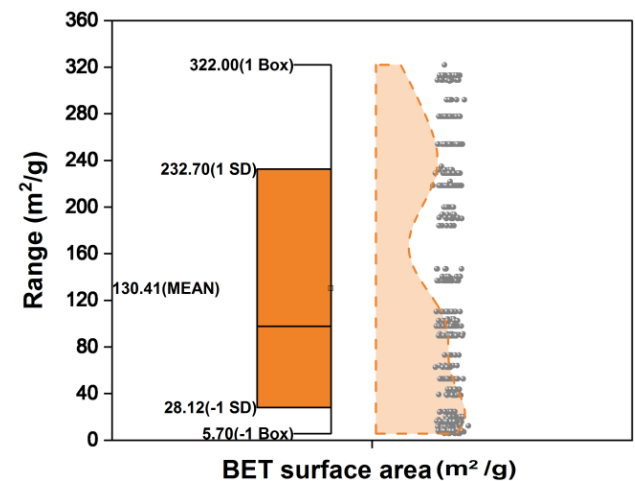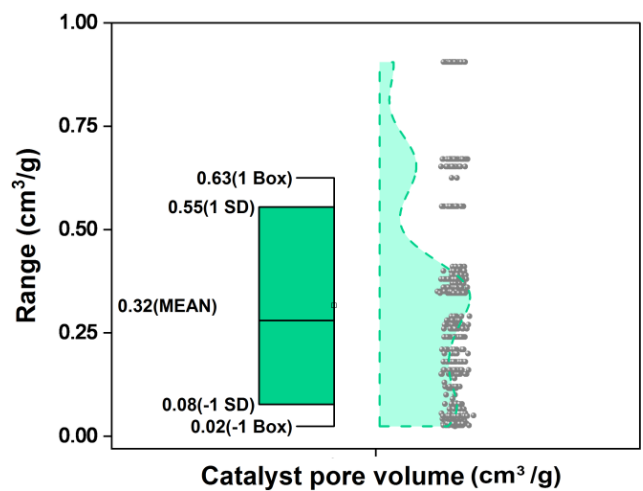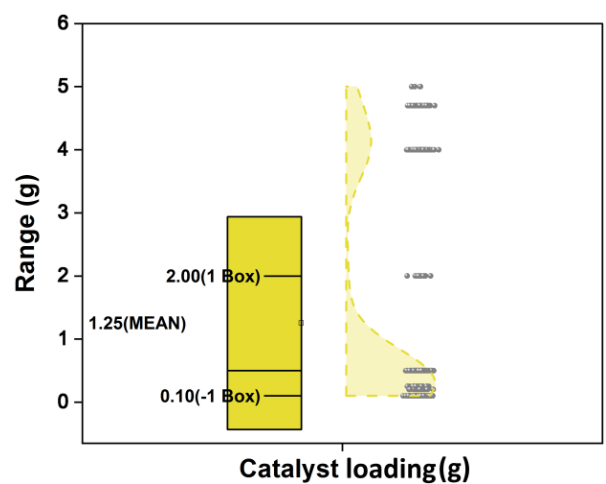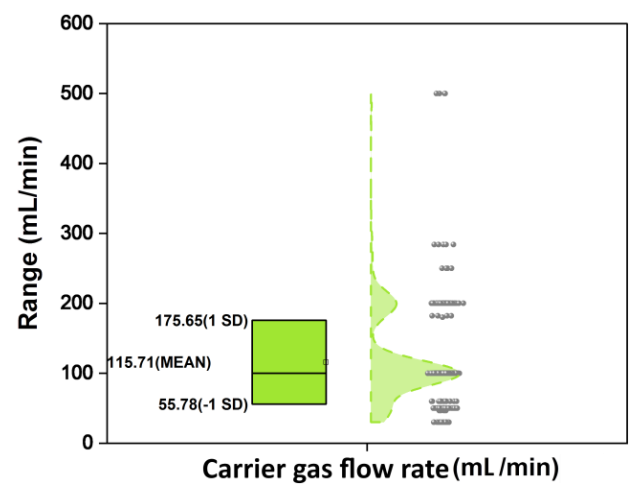

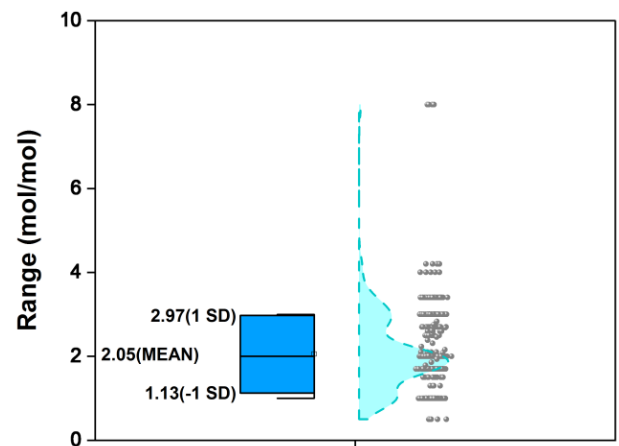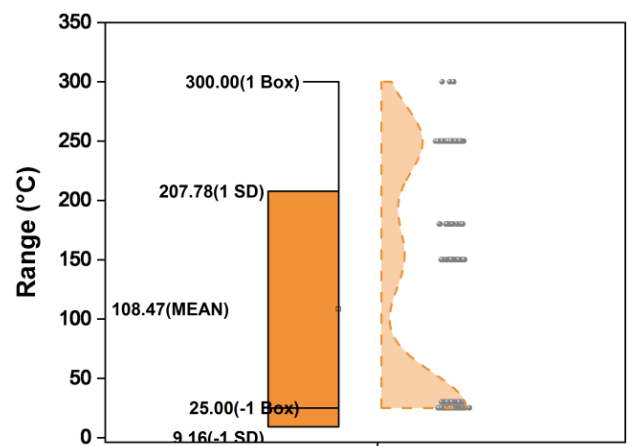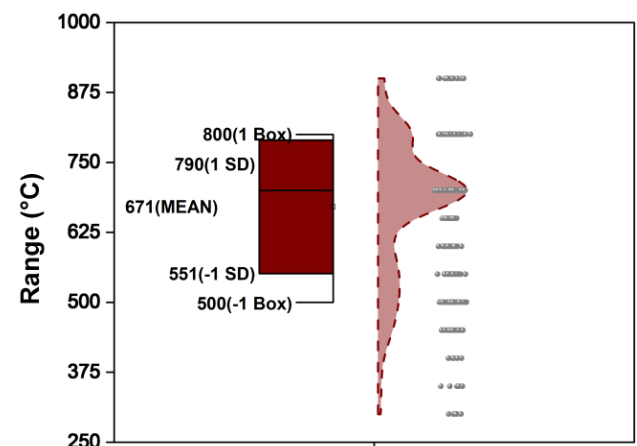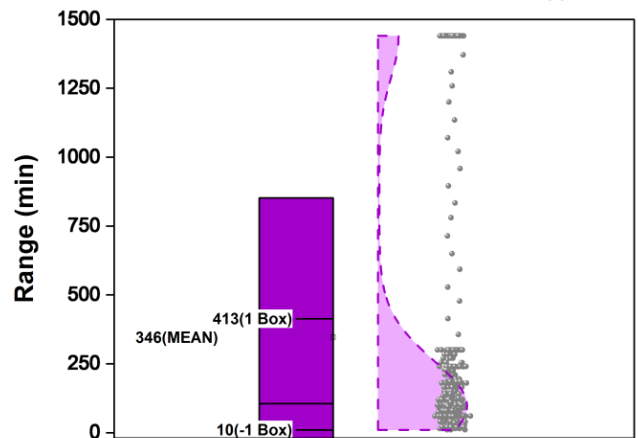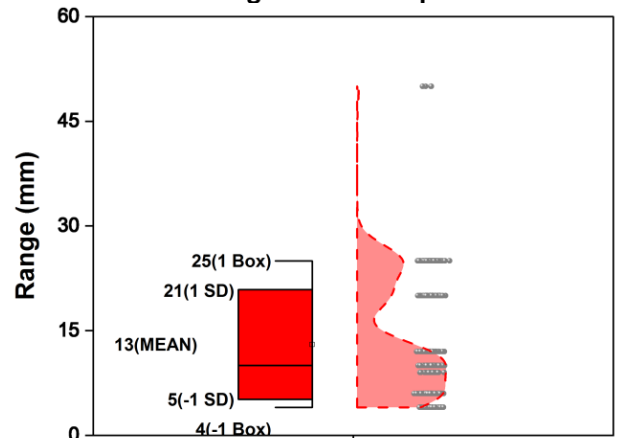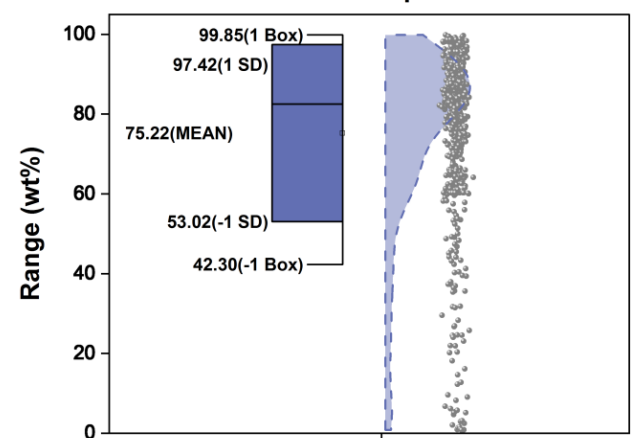

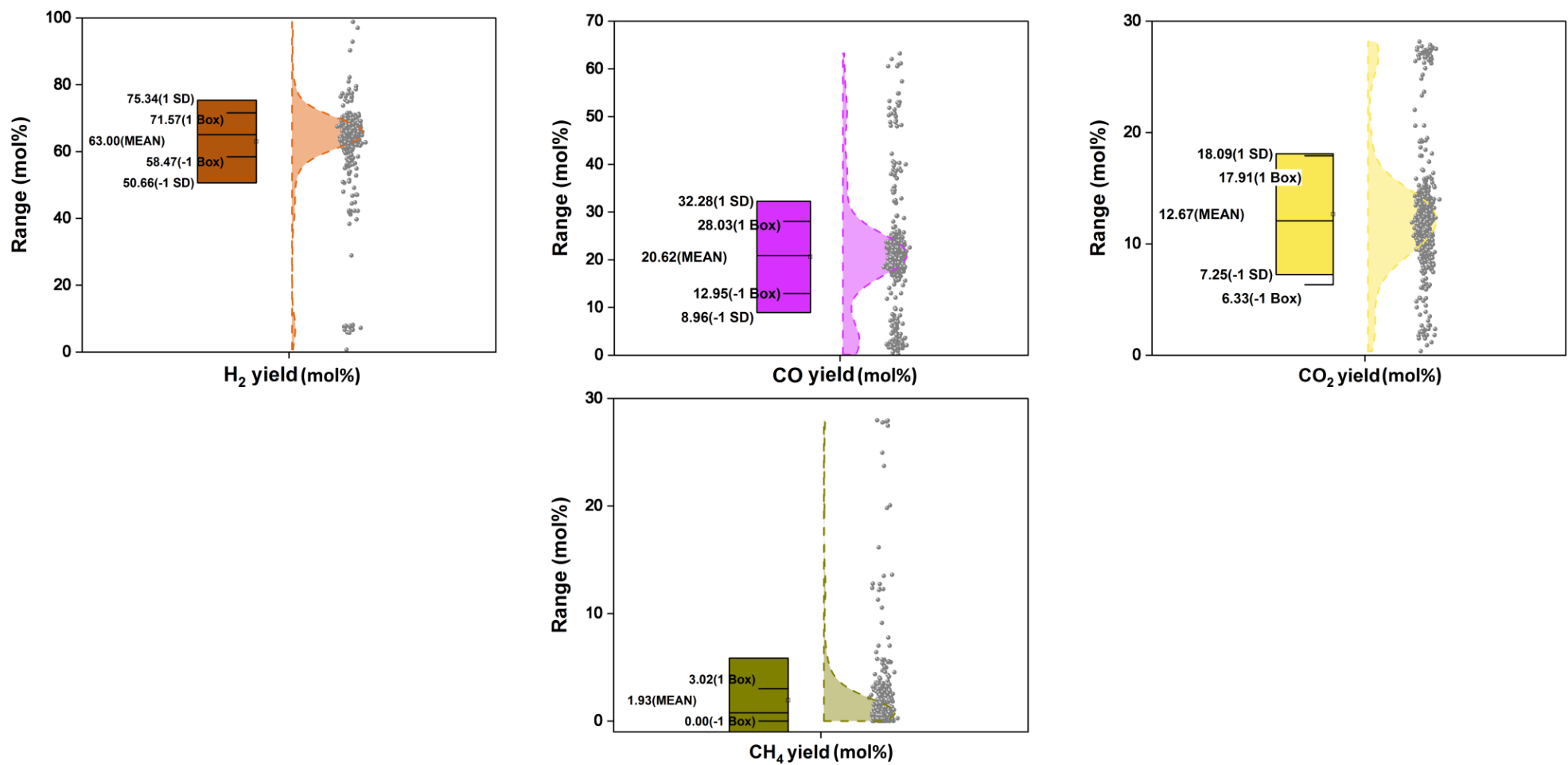

Figure 5. Descriptive statistics of the input/output parameters in the compiled database.

A basic statistical analysis encompassing width (the diversity of various catalyst properties and operating conditions) and depth (the number of data points) does not suffice for the compiled dataset. Thus, additional assessments such as Spearman correlation analysis, joint distributions using kernel density estimation, principal component analysis (PCA), and counterplot analysis were carried out to complete descriptive analytics of the compiled database. Spearman correlation analysis assesses the strength and preliminary relationships between pairs of variables, such as independent descriptors and dependent responses. This study employed Spearman correlation analysis to determine the correlation between independent input variables and dependent output responses in tar catalytic steam reforming. Based on the monotonic function, this nonparametric analysis measures the rank values of those two variables to assess the degree and direction of the link between them (George et al., 2021). Correlation coefficients of +1, −1, and 0 indicate the strongest positive, strongest negative, and weak correlations between the pair of variables, respectively. The matrix of Spearman correlation with the dendrogram for all combinations of the input features and output targets in the process is depicted in Figure 6.

The reaction temperature is positively correlated with toluene conversion (correlation coefficient ($r$) is +0.33) since the higher temperatures can be attributed to the enhancement of its steam reforming and thermal cracking reactions (Gao et al., 2021a). Increasing temperature can accelerate the decomposition of toluene into complex aromatic and heterocyclic hydrocarbons, decrease the amount of solid residue, and increase the yield and selectivity of light hydrocarbons (Gao et al., 2022; Liu et al., 2019). The tar catalytic steam reforming process is generally conducted at higher temperatures, increasing energy consumption and operating costs. The steam-to-carbon molar ratio is also positively correlated with toluene conversion ($r$ = +0.3), while it is negatively correlated with CO concentration ($r$ = −0.35). Steam injection in the toluene catalytic steam reforming process decreases the formation of carbon deposition by promoting the water-gas shift reaction and reducing CO concentration (Liu et al., 2019). Furthermore, increasing the steam-to-carbon molar ratio can greatly reduce coke formation by facilitating the gasification of deposited carbon (Tan et al., 2020b). It should also be noted that higher steam injection rates in the mixture can increase the formation of $CH_4$ by promoting the methanation reaction (Yue et al., 2010). It is worth noting that the steam-to-carbon molar ratio and toluene conversion may not always have a positive correlation and may even be slightly negative in some cases (Tan et al., 2020a). However, in this study, a positive correlation is observed from all the data extracted, suggesting that the overall impact of the steam-to-carbon ratio on the process is likely to be positive.

The BET surface area, pore volume, and crystallinity index are positively correlated with $H_2$ composition in syngas with $r = +0.33$, $r = +0.28$, and $r = +0.44$, respectively. Therefore, catalysts with higher BET surface areas and crystallinity index values could promote $H_2$-generation reactions, such as water-gas shift reaction (Pal et al., 2018). Spearman correlation analysis reveals a negative correlation ($r = -0.3$) between pore volume and toluene. This finding can be attributed to the potential blockage of the pore structure caused by coke formation and carbon deposition. As a result, the access of tar compounds into the catalyst active sites within the pores may have decreased, leading to a decline in toluene conversion. The reaction time is also negatively correlated with toluene conversion ($r = -0.28$). Increasing reaction time can reduce the catalyst activity by promoting pore blockage, coke deposition, and sintering over the catalyst (Artetxe et al., 2016; Ren et al., 2021). It should be noted that proper selection of reaction time can facilitate the tar catalytic steam reforming process by providing sufficient apparent activation energy required to perform reforming reactions (Gai et al., 2015; Yang et al., 2021)

To quantitatively understand the relationship and provide simple instructions in selecting proper conditions in toluene catalytic steam reforming, joint distributions using kernel density estimation for important features and output targets were explored (Figure 7). As shown in Figure 7, higher toluene conversion rates (higher than 80%) are achieved when the reaction temperature is above 650°C and the reaction time is lower than 200 min. As for reaction temperature, the value ranging from 650 to 750 °C is the optimal threshold for maximizing the $H_2$ concentration in the evolved syngas. This result could be obtained because of the promotion of water-gas shift reaction at this temperature range (Qian and Kumar, 2017). In addition, the steam-to-carbon molar ratio between 1.5–2.5 is the most favorable condition for maximizing toluene conversion and producing $H_2$-rich syngas.

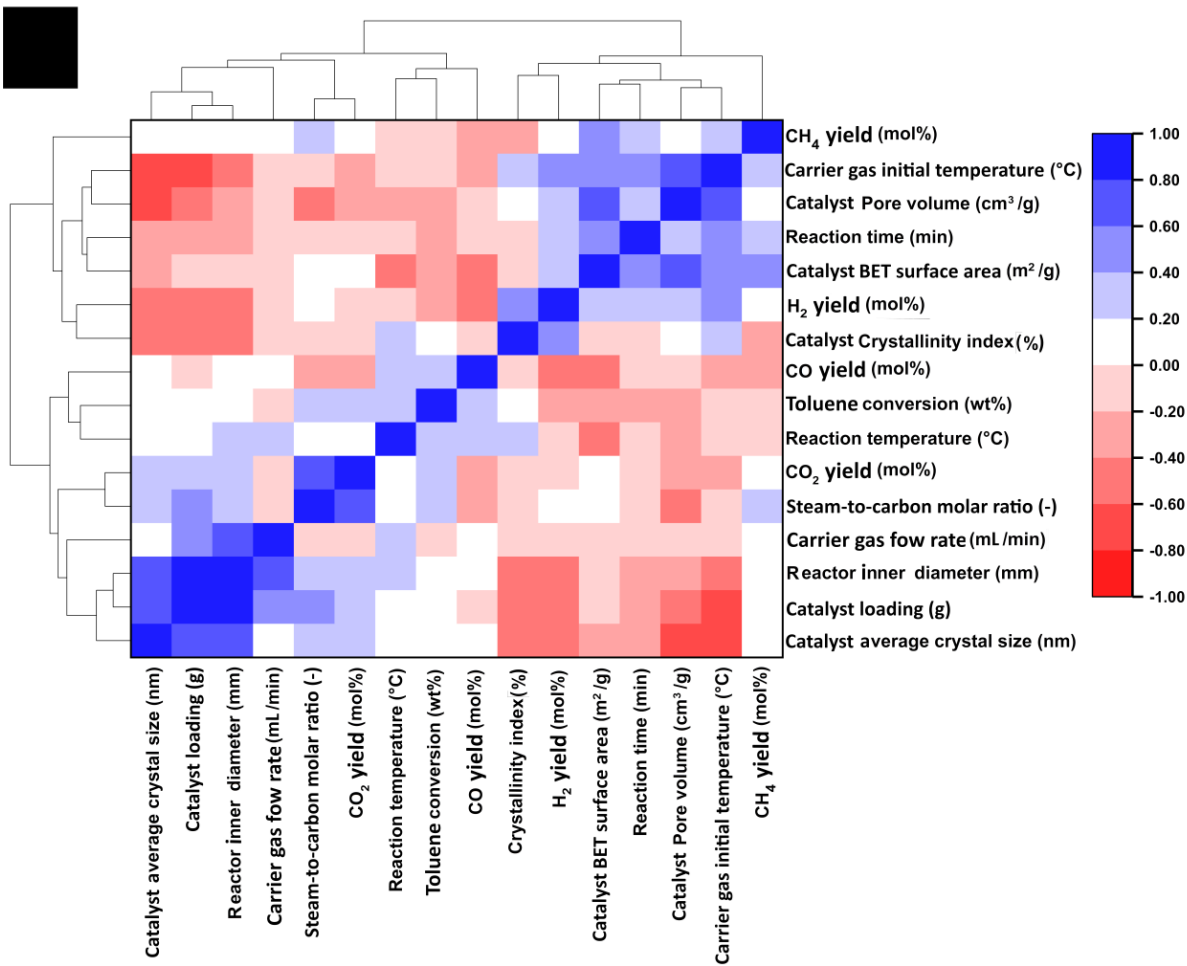

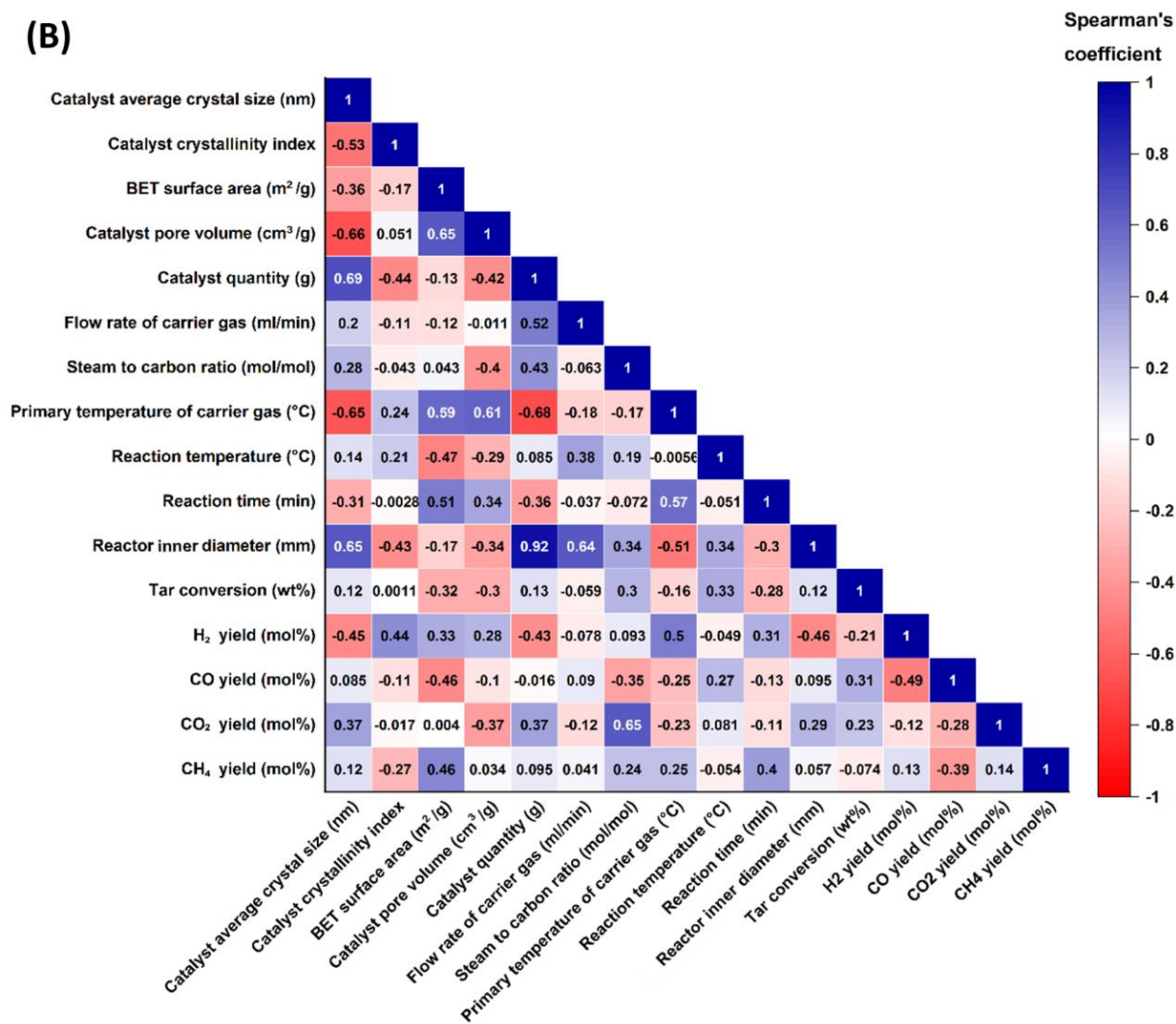

Figure 6. Dendrogram and hierarchical clustering heatmap (using Euclidean distance) for all combinations of input/output parameters based on the Spearman correlation matrix (A). The color bar indicates the correlation coefficient () among input/output variables (B), where blue and red represent positive and negative relationships, respectively.

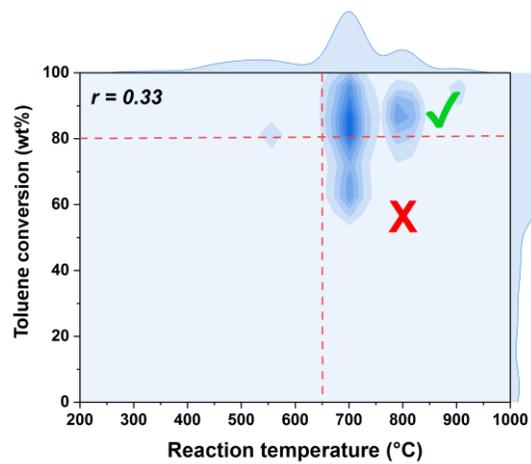 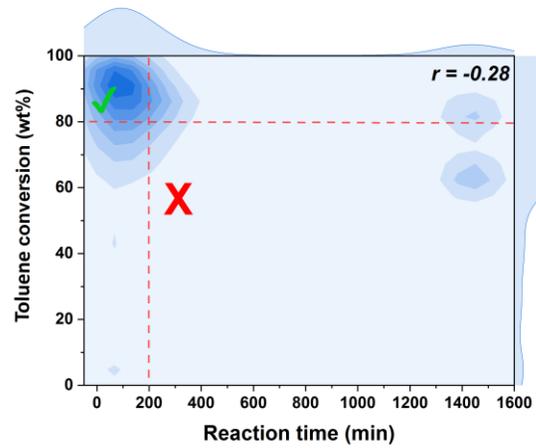 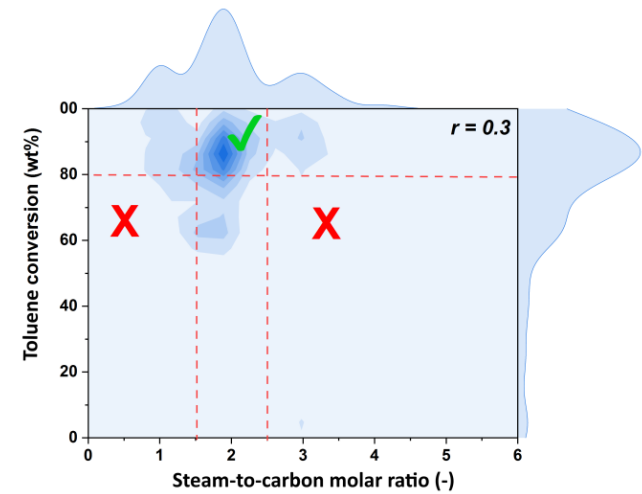
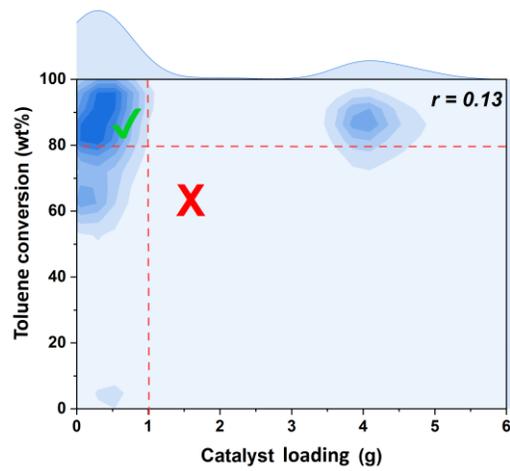 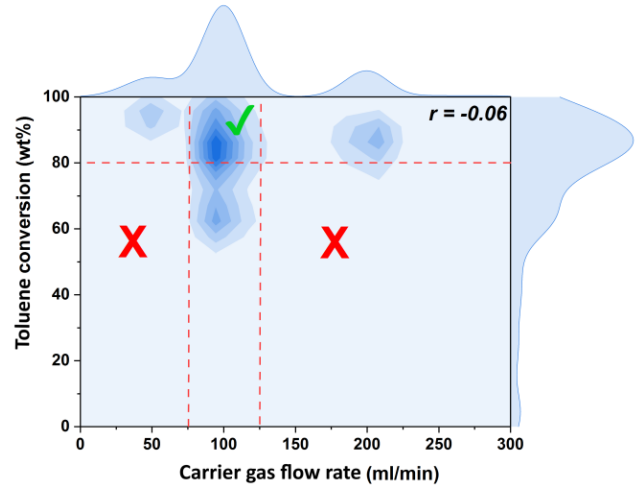 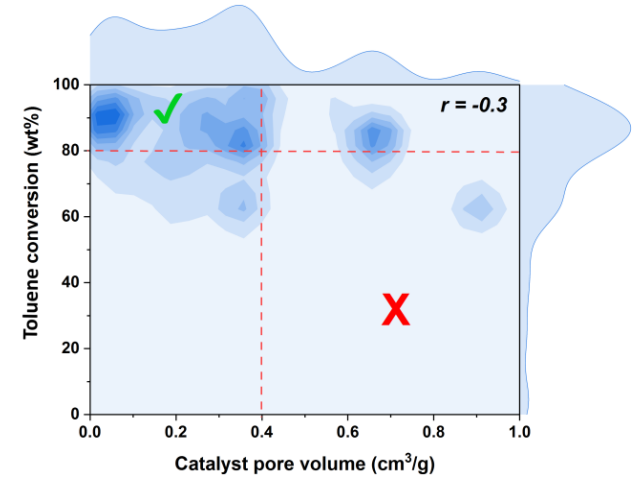

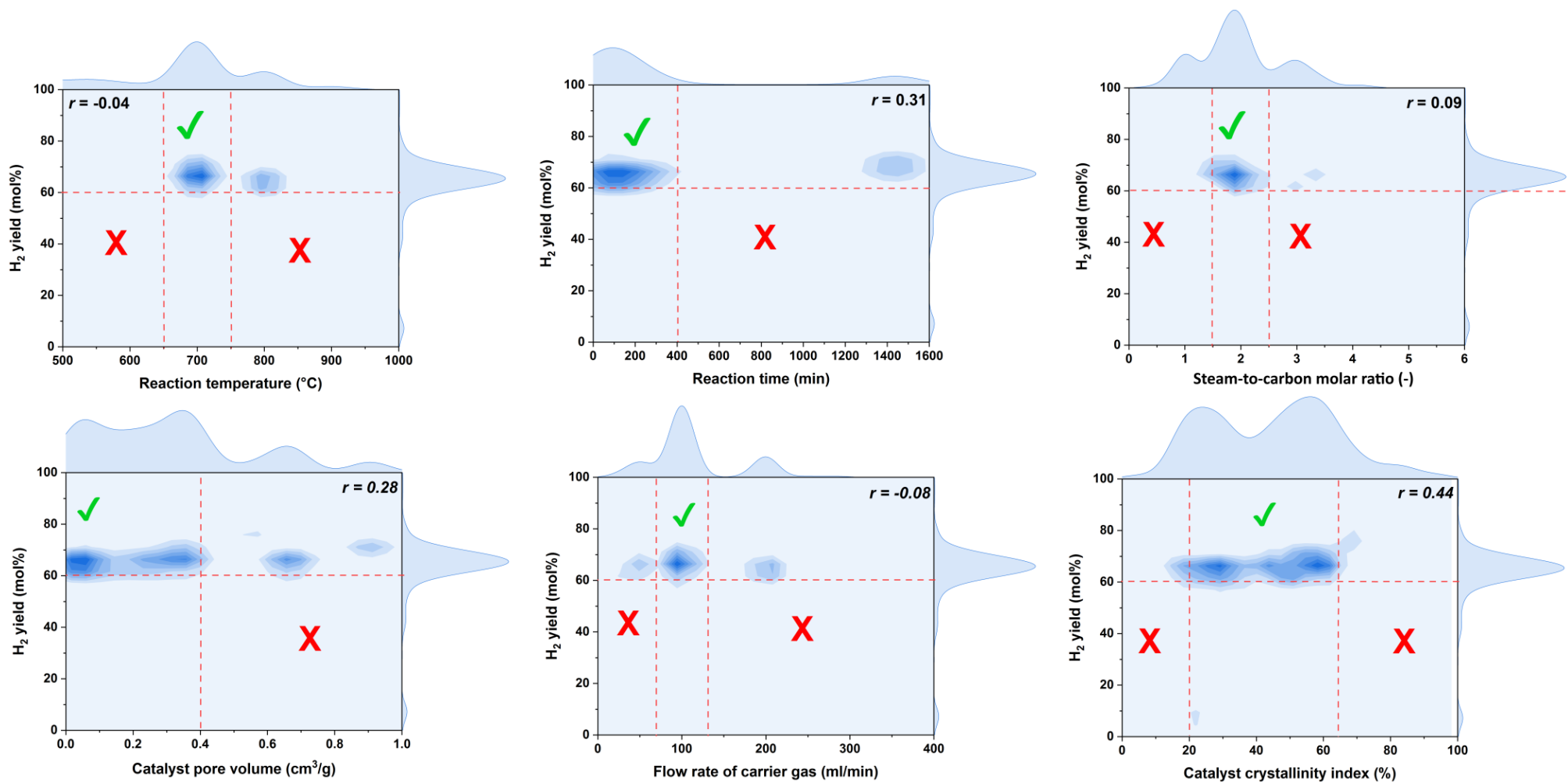

Figure 7. Evaluating the joint distributions using kernel density estimation for important input features and output targets. The marginal distributions are for the individual variables. The smooth kernel function is used to calculate bandwidth.

The PCA was conducted to understand the relationships between input variables and output responses in a given database. This analysis shows the relative positive/negative importance of each input variable on the desired responses before modeling. The first three principal components (PCs) account for more than 60% of the total variance, indicating the ability of these factors to effectively reflect the vast majority of information (Figure 8A). The first PC is mainly related to catalyst descriptors (average crystal size and pore volume), while the second and third PCs are correlated with reaction operating parameters (Figure 8B). The projection of the catalyst descriptors and operating parameters on the first three PCs in the database are illustrated in Figure 8C and Figure 8D, respectively. The correlation analysis indicates that the input features are significantly correlated with the tar conversion rate and the composition of the syngas. More specifically, the crystallinity index of the catalyst could positively affect toluene conversion. Thus, relatively higher quantities of crystalline materials (lower amorphous regions) could lead to higher catalyst activity and lower coke formation (Gai et al., 2019). The reaction temperature is positively correlated with the toluene conversion rate, indicating that higher reaction temperatures accelerate secondary reactions (e.g., cracking, reforming, and dehydrogenation). The promotion of secondary reactions could result in the decomposition of toluene into complex hydrocarbons (e.g., aromatics and olefins). Furthermore, increasing the temperature could enhance the endothermic steam reforming reactions of hydrocarbons, resulting in a significant rise in the CO and $H_2$ concentrations in the evolved syngas (Shen and Fu, 2018). It should be noted that these relationships may vary depending on the catalyst properties and operating conditions. Machine learning technology is applied in the following sections to comprehensively and systematically understand the relationships between the input features and the relative importance of independent features on the dependent responses in the tar catalytic steam reforming process.

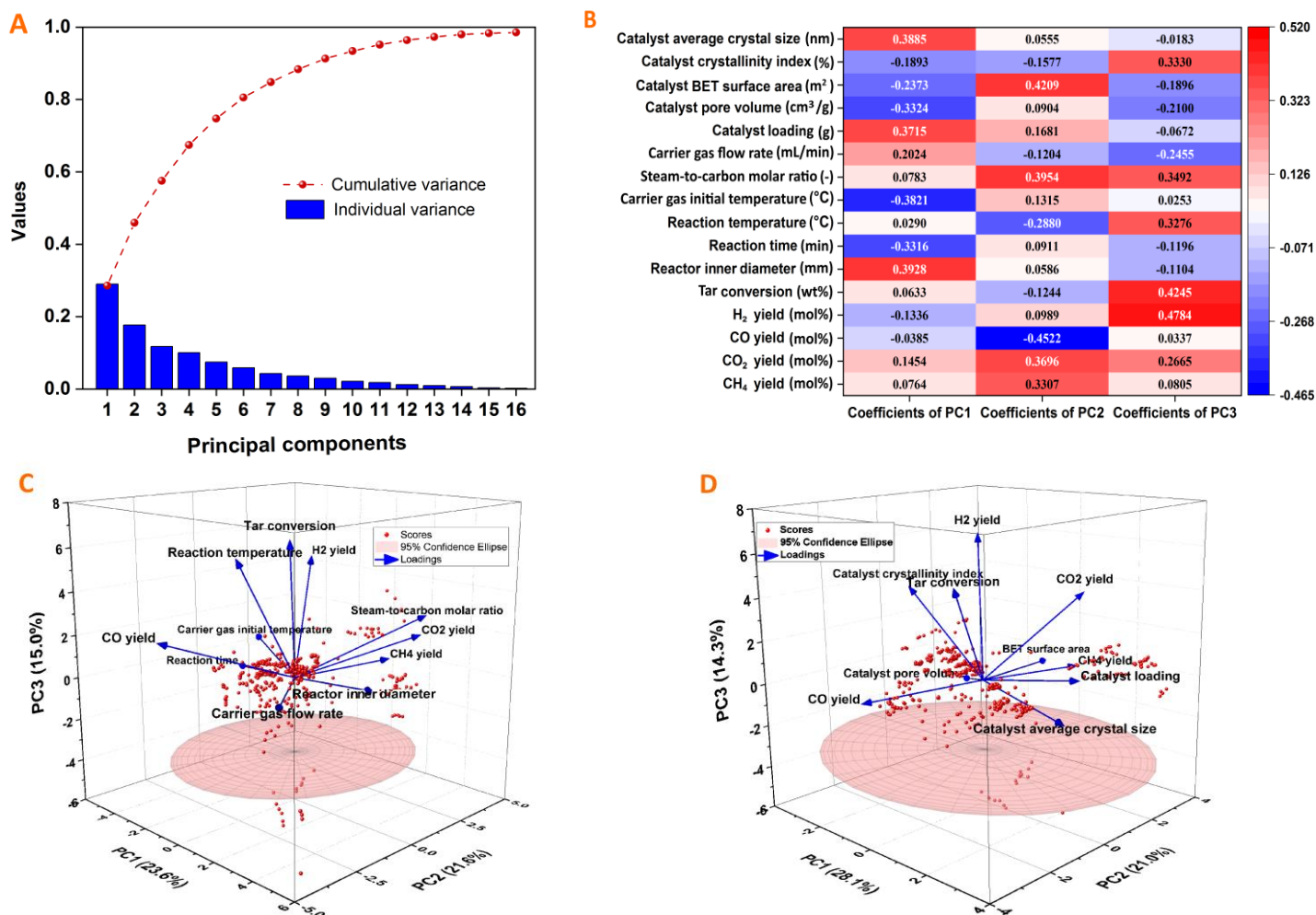

Figure 8. PCA of the compiled database: Variance of each component (A), correlation of independent inputs with the top-three components (B), effects of operating conditions on the output variables (C), and effects of catalyst descriptors on the output variables (D).

The relationship between the most influential operating parameters of toluene catalytic steam reforming (i.e., reaction temperature, steam-to-carbon molar ratio, and reaction time) and syngas composition can be shown using the contour diagram (Figure 9). The $H_2$-rich syngas is obtained when reaction temperature, steam-to-carbon molar ratio, and reaction time range 550–700 °C, 4.3–5.4 mol/mol, and 30–370 min, respectively. Optimizing the reaction temperature is crucial in tar catalytic steam reforming to obtain hydrogen-rich syngas. However, excessively high processing temperatures (i.e., > 900 °C) could potentially hinder the water-gas shift reaction (Gao et al., 2021a). This issue, in turn, results in a reduction in $H_2$ concentration in the evolved syngas. Furthermore, increasing reaction temperature can steadily decrease the concentrations of light hydrocarbons (e.g., $CH_4$ and $C_2H_6$) by promoting the endothermic steam reforming of tar (Laobuthee et al., 2015). The produced CO can further react with steam (water-gas shift reaction) at higher steam-to-carbon molar ratios, generating more $H_2$ (Guan et al., 2016). It should be noted that extremely high temperatures during toluene steam reforming can lead to rapid and irreversible deactivation of the catalysts used (Barbarias et al., 2016).

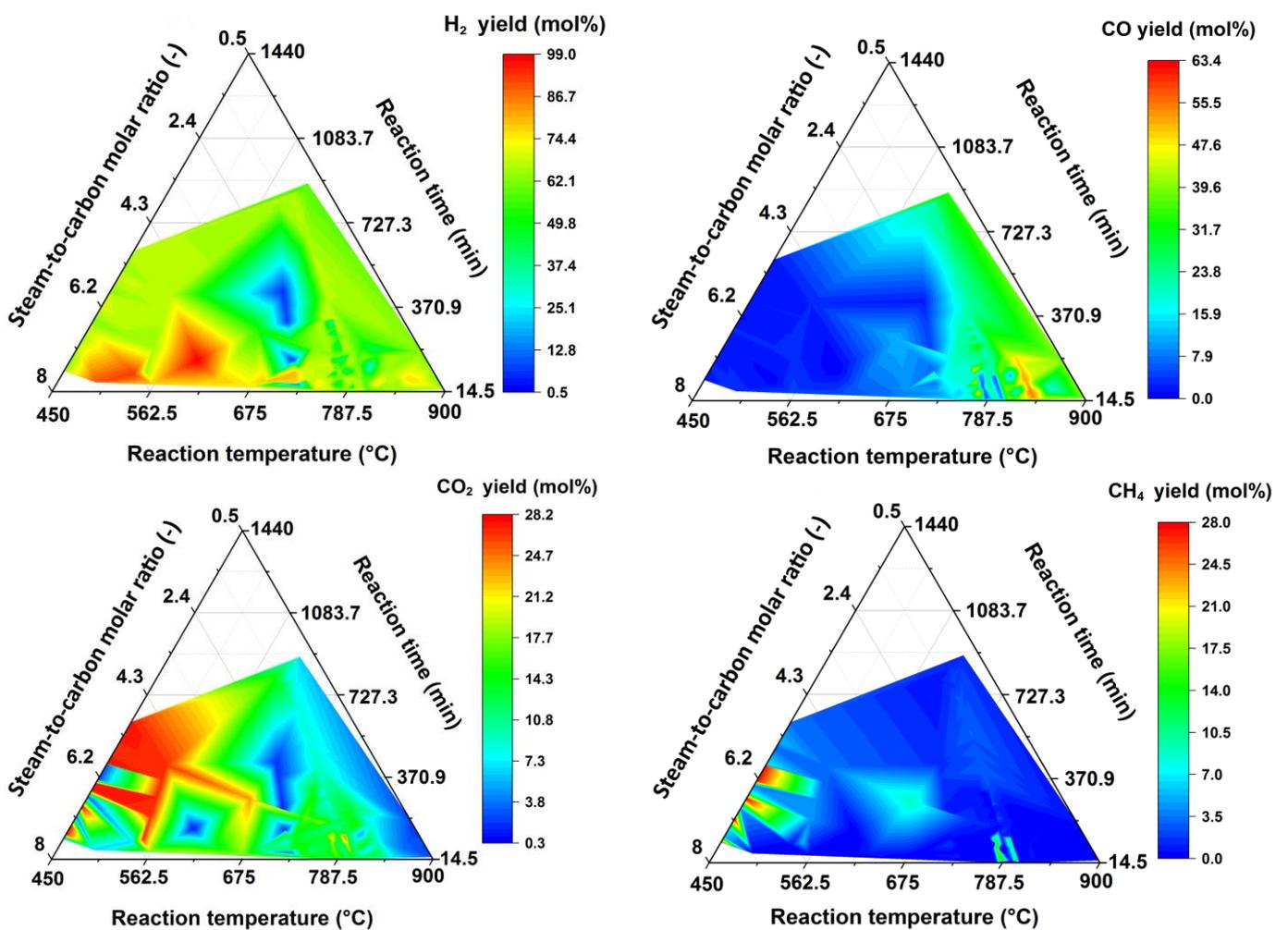

Figure 9. Effect of the main operating parameters on syngas composition during tar catalytic steam reforming.

### 3.2. Machine learning modeling results

The training and testing performance parameters ($R^2$, MAE, and RMSE scores) of the developed models in the k-fold cross-validation are shown in Figure 10. The EML algorithm provides the most accurate predictions for both training and testing datasets (maximum $R^2$ value and minimum MAE and RSME scores). The mean $R^2$, MAE, and RMSE scores of the developed EML models in the training phase for different outputs are between 0.992–0.998, $1\times10^{-3}$–$6.29\times10^{-5}$, and 0.007–0.037, respectively. These values are between 0.976–0.997, $1.19\times10^{-3}$–$2.85\times10^{-3}$, and 0.007–0.010 in testing, respectively. The calculated errors are generally satisfactorily low, and at the same time, the correlation coefficient is significant and close to unity. The high $R^2$ values and low MAE and RMSE values suggest that the EML approach is competent in predicting the desired responses of toluene catalytic steam reforming as a function of reaction conditions and catalyst properties.

According to Figure 10, the accuracy of the GPR model is close to the EML model. More specifically, the violin chart of the $R^2$, MAE, and RMSE scores for toluene conversion and $H_2$ yield is very close in both EML and GPR models. Nevertheless, EML performs slightly better than the GPR model in predicting CO, $CO_2$, and $CH_4$ yield. For both EML and GPR models, the statistical values are distributed tightly around the mean value for both the training and testing datasets. In fact, the EML method (multiply algorithm model) has proven to be strongly effective in the domains that are difficult to address using a single model-based system. EML creates several classifiers with relatively fixed (or similar) bias during the learning process and then combines their outputs by averaging to reduce the variance. This algorithm makes a trade-off relationship between bias (the accuracy of the classifier) and variance (the precision of the classifier) (Polikar, 2012). The outstanding results of the GPR model could be because of its great ability to solve nonlinear and nonparametric regression problems. It also shows good potential for interpolating between data points scattered in high-dimensional input spaces (Deringer et al., 2021).

The other four modeling methods do not show satisfying performance, particularly ANFIS, SVM, and GAM models. The violin plots related to these models do not indicate monotonous and stable predictions for the outputs considered. These models show a wide data distribution around the mean value for the $R^2$ score. In addition, a large bandwidth of density for the $R^2$ score can be seen in the plots. The same trend is observed for error scores. This issue suggests that the individual machine learning algorithms cannot precisely predict the complex toluene catalytic steam reforming process. Unlike these models, the EML model can accurately and systematically establish the relationships between the inputs and outputs of the process because of its unique feature of using multiple learning algorithms.

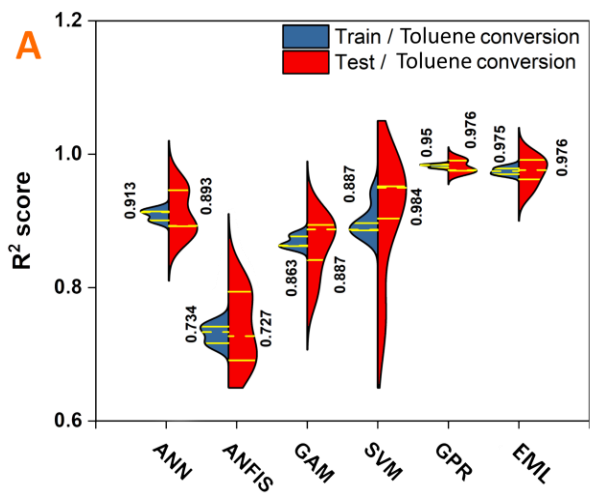 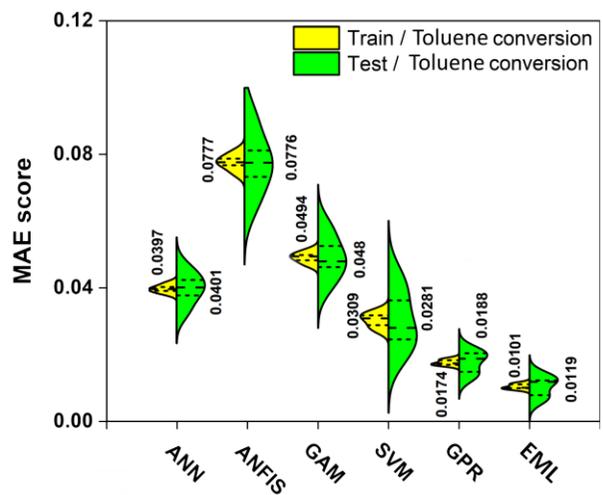 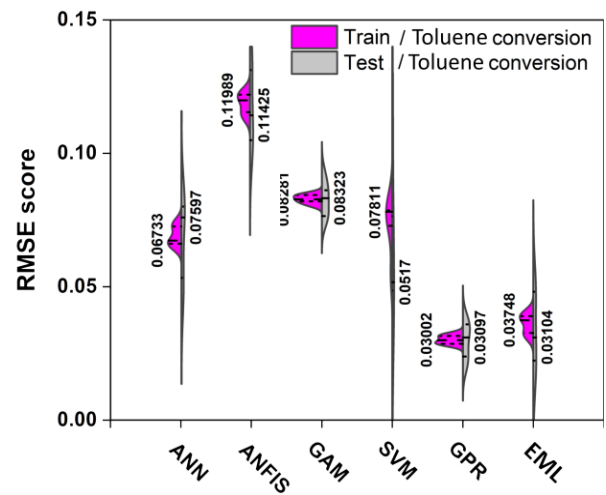

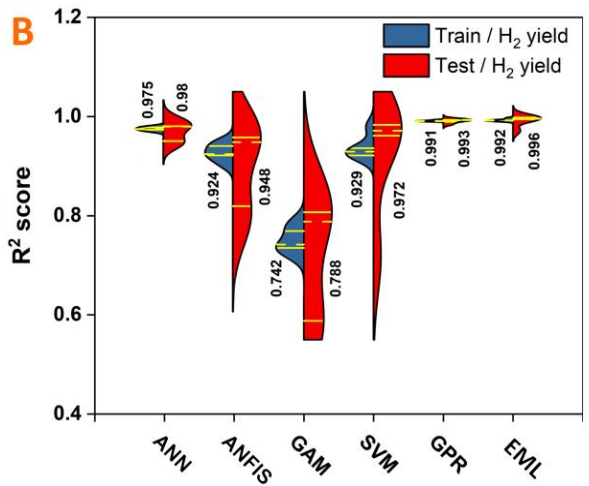 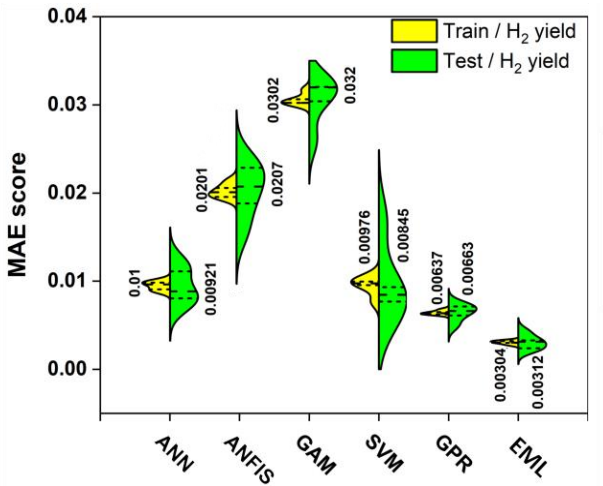 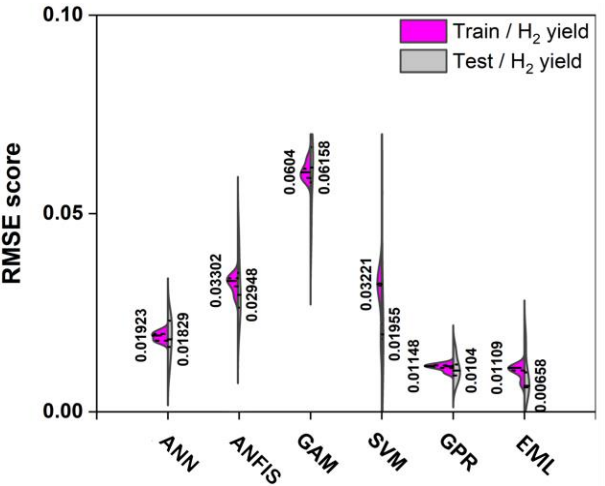

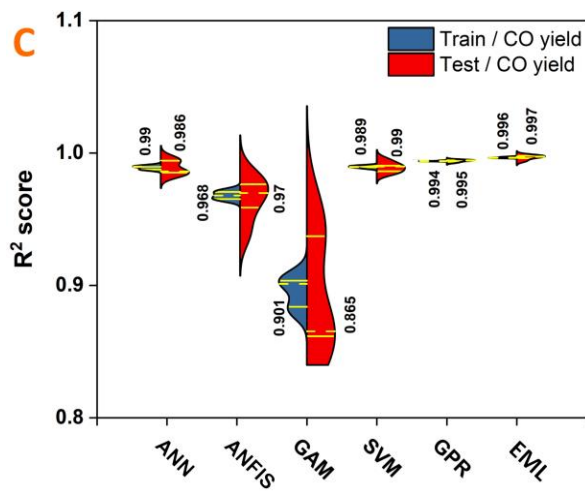 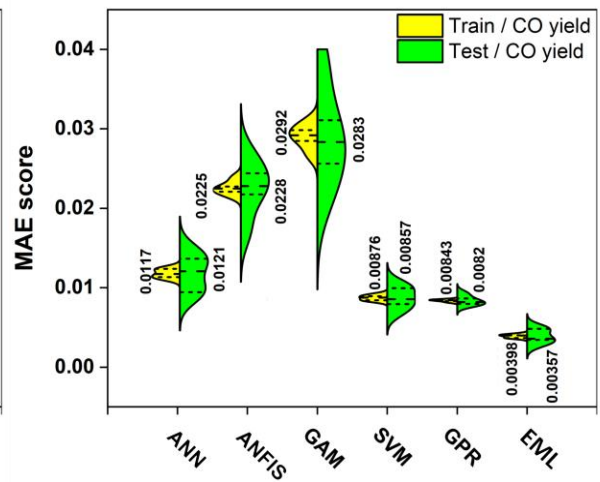 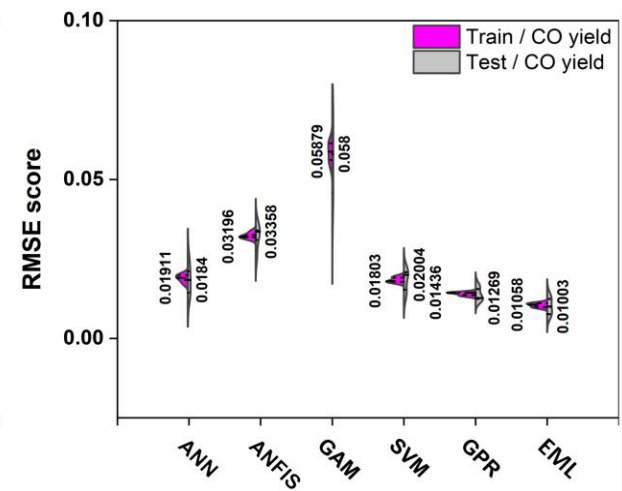
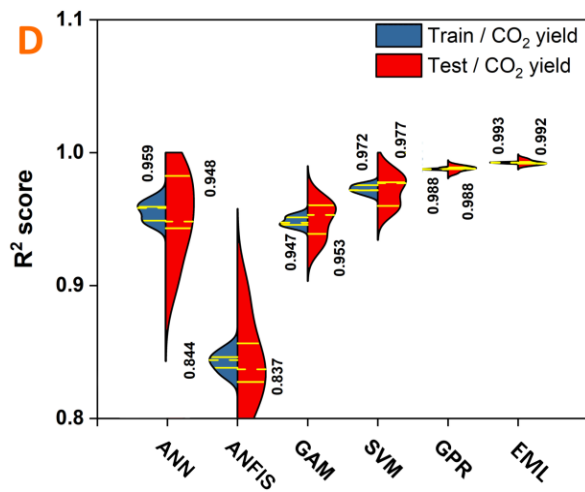 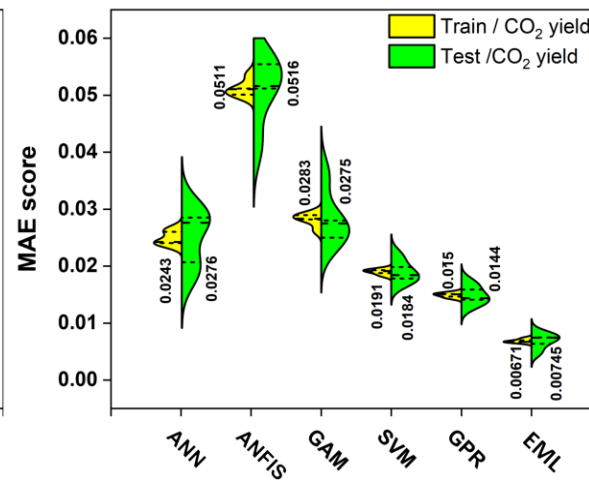 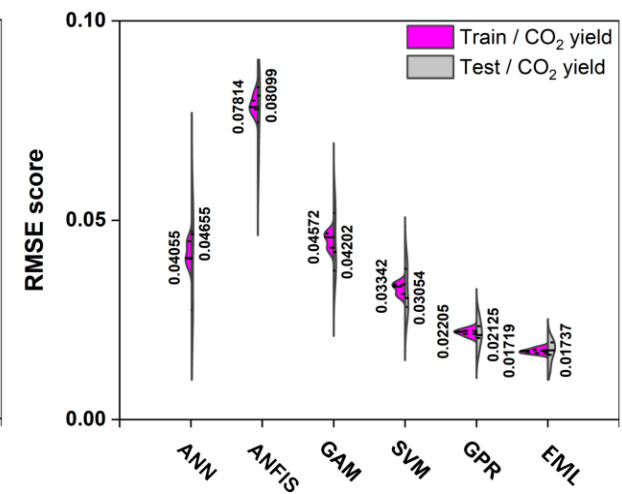

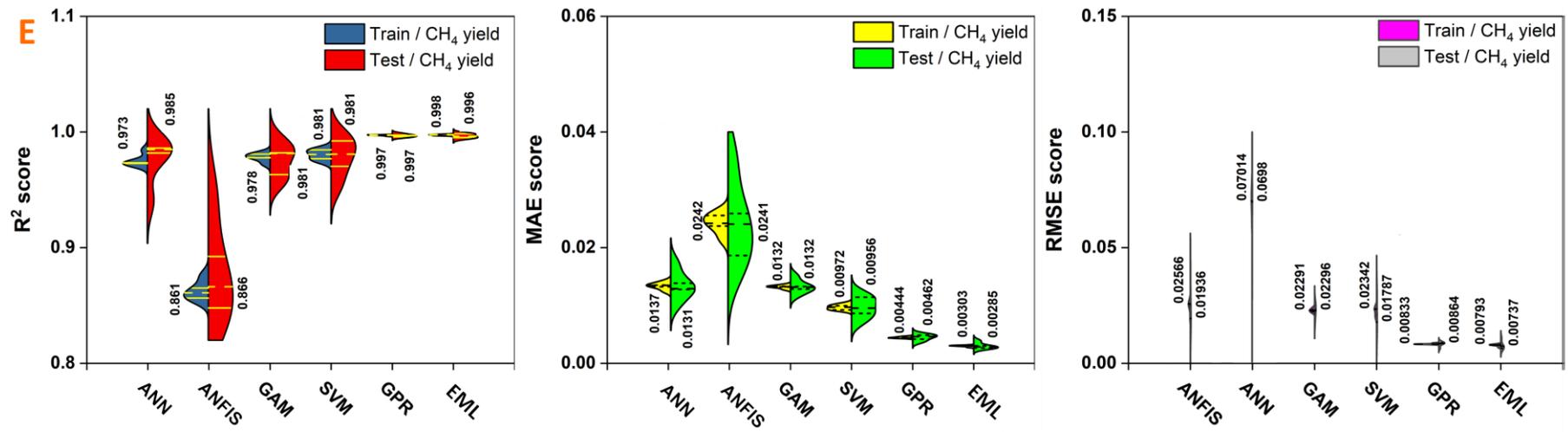

Figure 10. Training and testing performance parameters of the developed models in the k-fold cross-validation presented by violin charts for tar conversion (A), $H_2$ yield (B), CO yield (C), $CO_2$ yield (D), and $CH_4$ yield (E).

The predicted value against the experimental data with a coefficient shadow band of 95% (blue for training and red for testing phase) for all outputs in the 5-fold cross-validation is depicted in the Supplementary Word File (Figure S1). The dashed line in the figure depicts a 45-degree line, and the data are tightly distributed around the dashed line for the EML model. Obviously, the EML-predicted values have a positive linear correlation with experimental literature data. In addition, the $R^2$ scores for all the dependent responses in training and testing are very close using the EML model (Figure S1). This finding implies no overfitting in the EML model; thus, it was appropriately generalized. It should be noted that the performance of the GPR model is very close to the EML model. This issue might be because of the strong anti-noise characteristic of the GPR model when handling high-dimension data (Shafizadeh et al., 2022). The other four machine learning models are not close enough to the EML model. These models have many more points deviating from the experimental literature data than the EML and GPR models. The main reason for this finding could be the complexity and nonlinearity of toluene catalytic steam reforming. The EML could perform best when the least-squares boosting method is applied in the model. The maximum number of decision splits, minimum observation per leaf, number of ensemble learning cycles, and learning rate of the EML model were 6, 5, 250, and 0.295, respectively.

A user-friendly software platform was developed based on the MATLAB computer program to present the EML model in the simplest way (Supplementary TAR.exe File). This interface facilitates understanding and optimizing toluene catalytic steam reforming in future research attempts. It should be emphasized that the interface is intended to assist experimental measurements and not fully replace them. The user can use the developed interface to reduce costs and time by searching for the best catalysts and reaction conditions. The software can be run on computers without having MATLAB software already installed. Toluene conversion rate and syngas composition during catalytic steam reforming can be estimated by entering catalyst characteristics and reaction conditions (Figure 11).

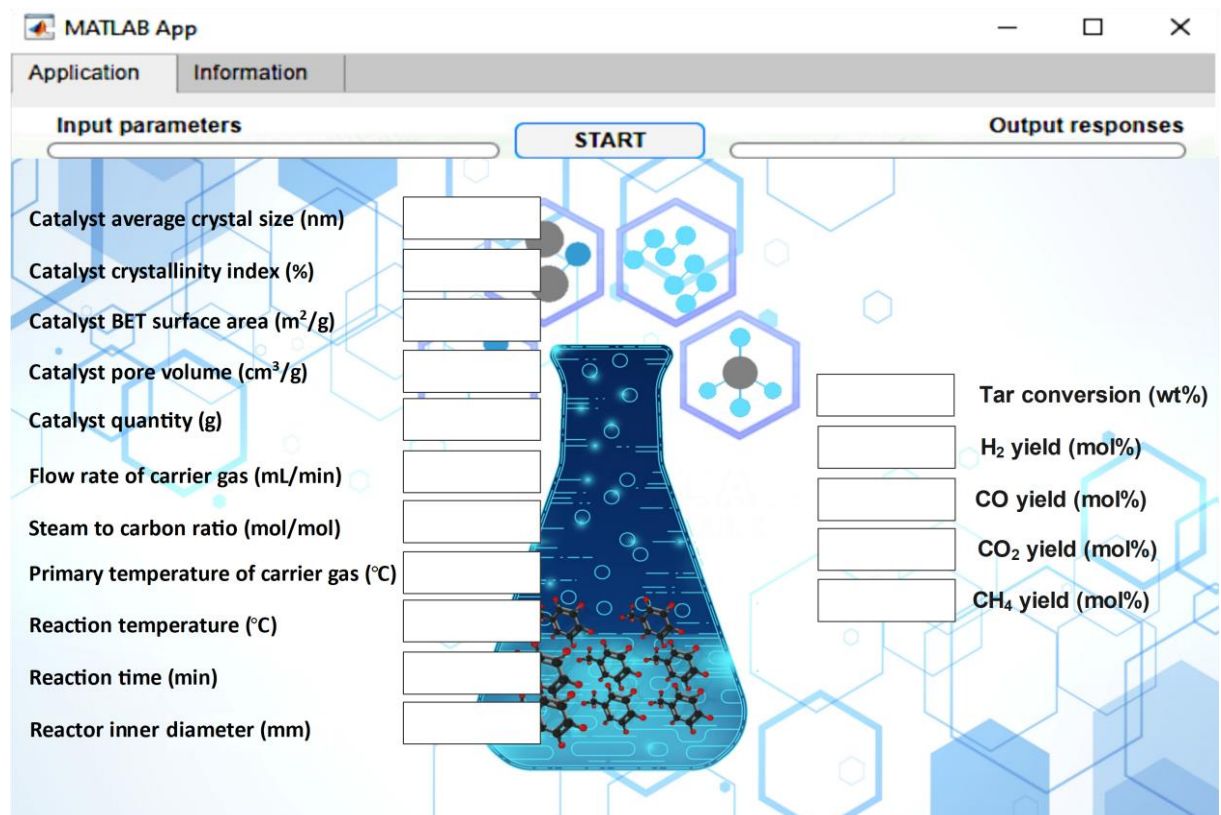

Figure 11. A screenshot of the developed Tar-APP.

## 3.4. Process optimization

The optimal catalyst characteristics and reaction conditions are provided in "Supplementary Excel File". Toluene conversion rate and syngas composition corresponding to the optimal points are also provided in "Supplementary Excel File". Table 1 also summarizes the range of optimal input parameters and corresponding outputs, including maximum tar conversion and $H_2$ yield, as well as minimum CO, $CO_2$, and $CH_4$ yield. The optimal catalyst characteristics and reaction conditions obtained through the MOPSO algorithm can be applied to laboratory-scale tar conversion systems, and the findings can serve as a foundation for developing large-scale reactors. These research endeavors pave the way for utilizing optimal operating parameters and catalyst characteristics to design and develop practical and efficient tar conversion systems.

Table 1. Summary result of the MOPSO algorithm to optimize the process.

| Input parameters | Average crystal size (nm) | Crystallinity index (%) | BET surface area (m$^2$/g) | Pore volume (cm$^3$/g) | Catalyst loading (g) | Carrier gas flow rate (mL/min) | Steam-to-carbon molar ratio (-) | Carrier gas initial temperature (℃) | Reaction temperature (℃) | Reaction time (min) | Reactor inner diameter (mm) |
|---|---|---|---|---|---|---|---|---|---|---|---|
| Optimal range | 11.52–16.09 | 17.22–29.65 | 476.03–638.55 | 0.373–0.469 | 3.30–3.99 | 289.38–378.25 | 5.81–7.15 | 273.93–279.58 | 637.44–725.62 | 599.7–879.49 | 19.12–25.05 |
| Output response | Toluene conversion (%) | | H$_2$ yield (mol%) | | CO yield (mol%) | | CO$_2$ yield (mol%) | | CH$_4$ yield (mol%) | | |
| Optimal results | 77.2–100 | | 88.74–94.01 | | 0–1.89 | | 0–1.36 | | 0.11–27.15 | | |

630

## 3.5. Feature importance analysis

Because of the black-box nature of machine learning models, recognizing the importance of input features and their impacts on output responses during the modeling process is challenging. Accordingly, SHAP analysis using the best-performing algorithm (EML model) was conducted to understand the effects of input descriptors on model prediction. This analysis can satisfactorily improve the interpretability of machine learning models by reflecting the important contribution value of each data point on the target features (i.e., local interpretation). In addition, the overall relevance of the input parameters (i.e., global pattern) can be provided by adding up the absolute SHAP values across the entire dataset (Li et al., 2021; Mu et al., 2022). This unique tool can aid in investigating nonlinear interactions of the features in toluene catalytic steam reforming based on cooperative game theory.

Figure 12 depicts the SHAP analysis of independent input features and the dependent output responses during toluene catalytic steam reforming. The order of features is defined as the maximum of absolute changes imposed on predicted values by modifying characteristics within their range. The left-hand graphs indicate the positive or negative effect of each specific input parameter with high (red dots) or low (blue dots) values on the corresponding output. A positive SHAP value indicates that the corresponding feature contributes to improving the prediction above the base value (SHAP value = 0). On the contrary, a negative SHAP value means that the corresponding feature contributes to the forecast being lower than the base value. The middle graphs depicted in Figure 12 illustrate the absolute maximum effect of each independent input parameter on the corresponding outputs. The right-hand pie charts in Figure 13 demonstrate the overall influence of catalyst properties and operating conditions on each output.

The reaction temperature is the most significant factor in tar conversion (Figure 12A). Higher temperatures generally accelerate endothermic reactions (e.g., decomposition and dehydrogenation of aromatic substances) involved in steam reforming, increasing tar conversion (Díez et al., 2020). However, excessive temperature (e.g., >800 °C, as shown in Figure 7) leads to the catalyst sintering phenomenon, decreasing tar conversion (Gu et al., 2020). The catalyst BET surface area, average crystal size, and crystallinity index are the next influential parameters in tar conversion. The impact of BET surface area on tar conversion is negative, i.e., a higher BET surface area results in a lower tar conversion. The increase in BET surface area is mainly attributed to the increased number of microspores (L. Ren et al., 2020). Generally, the microspores can quickly be blocked during the tar conversion process, causing viscous tar aggregation on the surface and lowering catalyst activity. Unlike reaction temperature, the SHAP values of BET surface area are significantly more moderate in both positive and negative directions, indicating that the effect on increasing and decreasing the prediction results was much lower.

The average crystal size of the catalyst is the next greatest contributor to the prediction of tar conversion. This feature does not show any noticeable positive or negative correlation during the modeling process, suggesting that the average crystal size of the catalyst may not play a decisive role in the conversion of tar compounds. Catalyst loading has a positive correlation with tar conversion. Increasing catalyst loading in tar reforming could increase the active catalytic sites provided (Karnjanakom et al., 2015). It should be noted that increasing catalyst loading beyond the optimal value could hinder the mass transfer phenomenon during the tar reforming process, thereby lowering catalytic activity (Quan et al., 2020). The reaction time is negatively correlated with tar conversion since a longer reaction time implies higher catalyst deactivation due to the coke deposition during the tar reforming process. Overall, the contribution of operating conditions (49.1%) and catalyst properties (50.9%) to tar conversion is very close.

Like tar conversion, the reaction temperature could contribute the most to the $H_2$ yield (Figure 12B). It should be noted that the concentration of $H_2$ is decreased by increasing the reaction temperature (e.g., > 750 °C, as shown in Figure 7). This occurs due to the promotion of the reverse water gas shift reaction, which is thermodynamically favorable at higher temperatures. Furthermore, the reverse water-gas shift reaction increases CO selectivity and decreases at such elevated temperatures (see Figure 12C-D) (Ashok and Kawi, 2015). Catalyst average crystal size has no noticeable positive and negative correlations with the SHAP values of the $H_2$ yield. Therefore, this input parameter does not play a decisive role in the $H_2$ yield. The carrier gas initial temperature has almost zero SHAP values, signifying the negligible impact on $H_2$ yield. The carrier gas should warm up to an optimal value before injecting into the reactor to avoid tar condensation. The steam-to-carbon molar ratio is among the top two most influential parameters, with a negative correlation in the yield of CO and $CO_2$. The concentration of CO is decreased by increasing the steam-to-carbon molar ratio because of the acceleration of the water-gas shift reaction. It should be noted that excessive steam (water) injection leads to adsorption saturation of steam on the catalyst surface, decreasing tar decomposition during the process (Zou et al., 2018). In addition, higher steam-to-carbon molar ratios (e.g., > 3 mol/mol, as shown in Figure 7) result in additional energy consumption for water evaporation, sintering possibility of the active catalyst sites, and significantly impress the economic viability of the process. The results show that the contribution of operating conditions to the yield of CO and $CO_2$ is more noteworthy than catalyst properties. However, operating conditions (50.3%) and catalyst properties (49.7%) have almost equal contributions to $CH_4$ yield (Figure 12E).

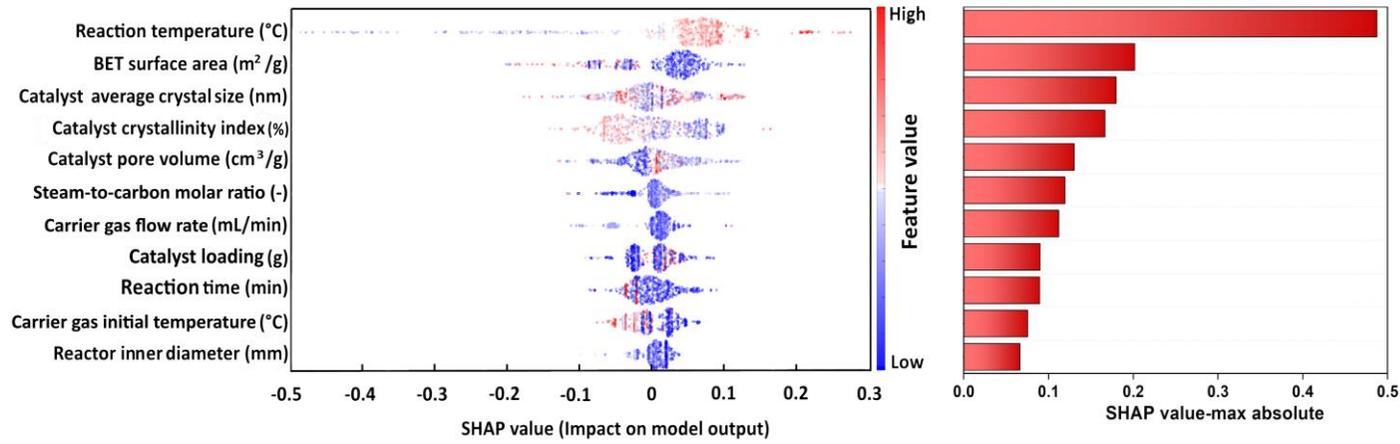
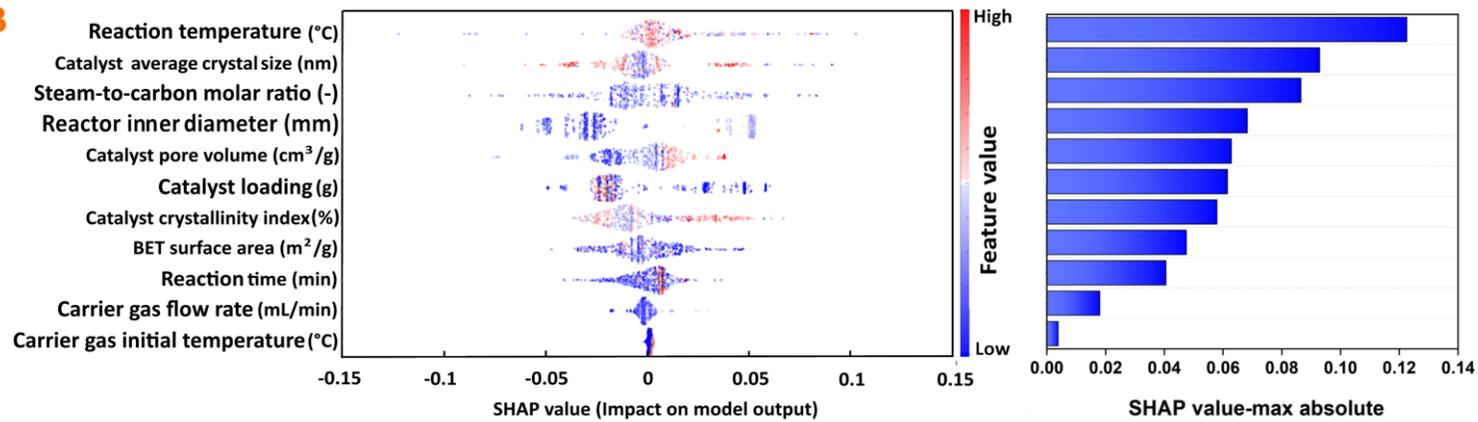

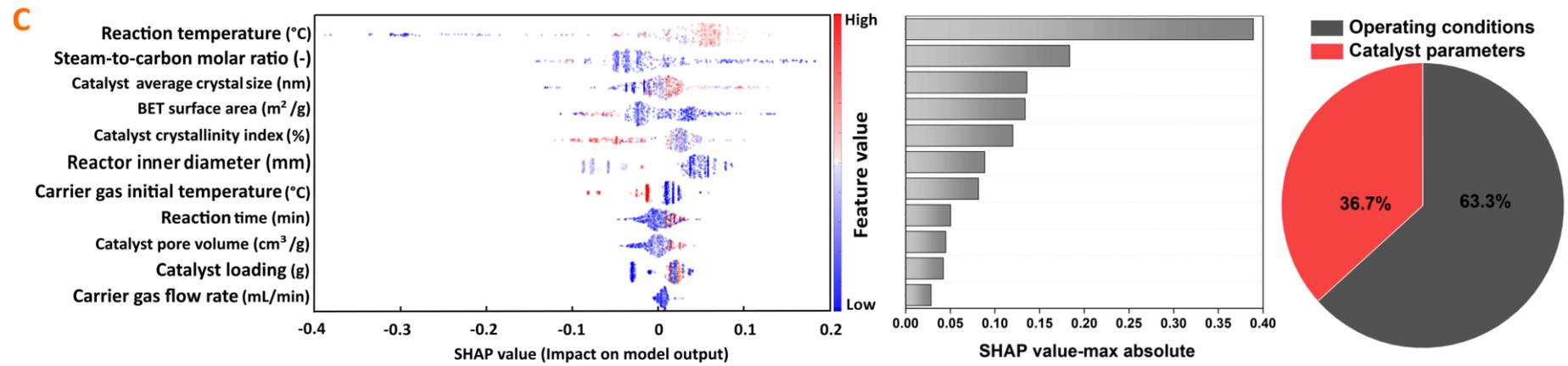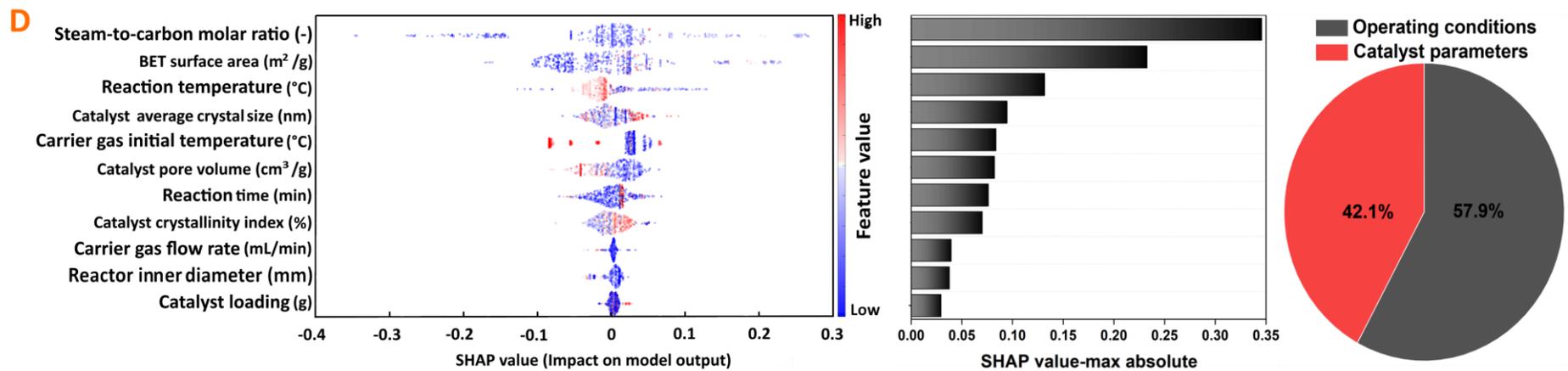

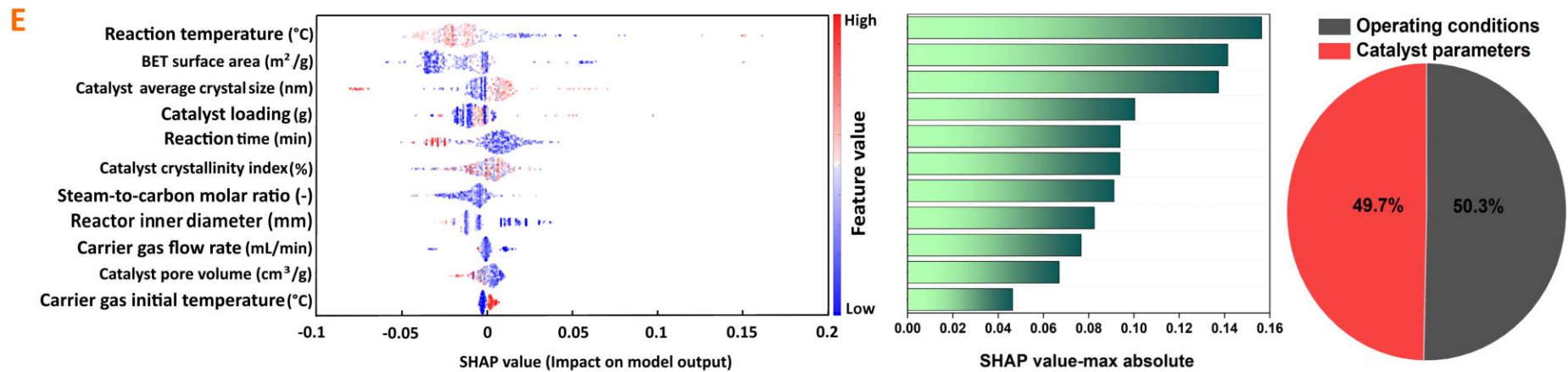

Figure 12. Effects of independent input parameters on tar conversion (A), $H_2$ yield (B), CO yield (C), $CO_2$ yield (D), and CH4 yield (E) obtained using the SHAP analysis.

Throughout this study, machine learning technology was applied to predict toluene catalytic steam reforming based on the available experimental data in the literature, with no assumptions and simplifications. Therefore, machine learning technology could provide better generalization ability for toluene catalytic steam reforming under various reaction conditions and catalyst characteristics. Nevertheless, model prediction accuracy depends on input data availability and quality. With the tailored development of machine learning models for understanding the toluene catalytic steam reforming process, the prediction ability of developed models can significantly decrease the time-consuming and cost-intensive experimental procedures for designing and testing catalysts and optimizing reaction conditions.

## 4. Concluding remarks and perspectives

An evolutionary machine learning framework was developed to model, understand, and optimize catalytic steam reforming of toluene as a tar model compound. 592 data patterns covering a wide range of catalyst characteristics (average crystal size, crystallinity index, BET surface area, and average pore volume) and reaction conditions (catalyst loading, carrier gas flow rate, steam-to-carbon molar ratio, carrier gas initial temperature, reaction temperature, processing time, and reactor inner diameter) were compiled from the published literature. Using an innovative approach, the crystallinity index and average crystal size of the catalyst were obtained from the XRD curves reported in the literature. It is worth mentioning that the method used to extract information from XRD graphs can be used in other research domains.

The compiled database was statistically analyzed to extract knowledge from experimental data on the catalytic steam reforming of toluene. Six machine learning models (ANN, ANFIS, GAM, SVM, GPR, and EML) were tuned by the PSO algorithm to predict toluene conversion rate and syngas composition based on catalyst characteristics and operating parameters. EML could provide the best prediction performance, with an $R^2 > 0.976$ and an MAE < $2.85×10^{-3}$. The MOPSO algorithm could satisfyingly seek optimal catalyst characteristics and reaction conditions by maximizing toluene conversion and $H_2$ yield while minimizing CO, $CO_2$, and $CH_4$ yield.

SHAP analysis was conducted to understand the effects of input descriptors on predicting desired responses. In general, the reaction temperature was the most important parameter in tar conversion and syngas composition, followed by the BET surface area and the average crystal size of the catalyst. A software platform was developed based on the EML model to accelerate catalyst design and reduce experimental costs. Overall, it could be concluded that machine learning technology could successfully extract useful information about tar catalytic steam reforming, even from datasets from various sources across different research laboratories.

Despite the promising features of the developed machine learning model, it can only be used to characterize catalytic steam reforming of toluene. Fortunately, the proposed generic workflow can be applied to other tar model compounds (i.e., benzene, phenol, methyl naphthalene, indene, anisole, and furfural) or even similar research fields. Future research should attempt to develop a comprehensive machine learning modeling covering a wide range of tar model compounds. Similar to XRD graphs, fundamental information about catalyst structures can be obtained from X-ray photoelectron spectroscopy graphs. Future work should focus on extracting information from these graphs to make more comprehensive and inclusive input descriptors. In addition, some other effective properties of heterogeneous catalysts (e.g., catalyst acid sites or lattice oxygen activity) can be considered in the modeling process to further enrich input descriptors.

Future machine learning modeling can focus on specific groups of catalysts (transitional or alkaline metal-based catalysts) or even more specific types of metal-based catalysts (i.e., nickel-based catalysts). In such a modeling scenario, other effective parameters, such as metal loading and metal/support/promoter types, can be input features. Besides the mentioned properties, the chemical properties of the catalyst can be included as input descriptors. For instance, some useful input features can be obtained from temperature-programmed desorption analysis. In addition, the steam reforming of real tar produced during the gasification or pyrolysis process should be considered in future attempts to develop universal machine learning models. The machine learning models presented in this study were developed based on a few catalyst types. Therefore, a more complete dataset covering more catalyst types and their characteristics is required to improve the universality of machine learning models in this domain.

**Acknowledgements**

J.Pan would like to acknowledge the financial support from the National Natural Science Foundation of China (42271401) and the Special funds for basic scientific research business of central public welfare research institutes (G2022-01-16). The authors would like to thank Universiti Malaysia Terengganu under International Partnership Research Grant (UMT/CRIM/2-2/2/23 (23), Vot 55302) for supporting this joint project with Henan Agricultural University under a Research Collaboration Agreement (RCA). This work is also supported by the Ministry of Higher Education, Malaysia under the Higher Institution Centre of Excellence (HICoE), Institute of Tropical Aquaculture and Fisheries (AKUATROP) program (Vot. No. 63933 & Vot. No. 56051, UMT/CRIM/2-2/5 Jilid 2 (10) and Vot. No. 56052, UMT/CRIM/2-2/5 Jilid 2 (11)). The manuscript is also supported by the Program for Innovative Research Team (in Science and Technology) in the University of Henan Province (No. 21IRTSTHN020) and Central Plain Scholar Funding Project of Henan Province (No. 212101510005). The authors would like to acknowledge that this work, in part, has been conducted under the umbrella of the MoU between the University of Saskatchewan (Canada) and Universiti Malaysia Terengganu (Malaysia). The authors would also like to extend their sincere appreciation to the University of Tehran and the Biofuel Research Team (BRTeam) for their support throughout this project.

Energy 14, 545–569. https://doi.org/10.1007/s11708-020-0800-2

Tanabe, K.K., Cohen, S.M., 2010. Modular, Active, and Robust Lewis Acid Catalysts Supported on a Metal−Organic Framework. Inorg. Chem. 49, 6766–6774. https://doi.org/10.1021/ic101125m

Umenweke, G.C., Afolabi, I.C., Epelle, E.I., Okolie, J.A., 2022. Machine learning methods for modeling conventional and hydrothermal gasification of waste biomass: A review. Bioresour. Technol. Reports 17, 100976. https://doi.org/10.1016/j.biteb.2022.100976

Üstün, İ., Üneş, F., Mert, İ., Karakuş, C., 2020. A comparative study of estimating solar radiation using machine learning approaches: DL, SMGRT, and ANFIS. Energy Sources, Part A Recover. Util. Environ. Eff. 1–24. https://doi.org/10.1080/15567036.2020.1781301

Vivanpatarakij, S., Assabumrungrat, S., 2013. Thermodynamic analysis of combined unit of biomass gasifier and tar steam reformer for hydrogen production and tar removal. Int. J. Hydrogen Energy 38, 3930–3936. https://doi.org/10.1016/j.ijhydene.2012.12.039

Wang, D., Tan, D., Liu, L., 2018. Particle swarm optimization algorithm: an overview. Soft Comput. 22, 387–408. https://doi.org/10.1007/s00500-016-2474-6

Wu, W., Fan, Q., Yi, B., Liu, B., Jiang, R., 2020. Catalytic characteristics of a Ni–MgO/HZSM-5 catalyst for steam reforming of toluene. RSC Adv. 10, 20872–20881. https://doi.org/10.1039/D0RA02403A

Wu, Z., Zhu, D., Chen, Z., Yao, S., Li, J., Gao, E., Wang, W., 2022. Enhanced energy efficiency and reduced nanoparticle emission on plasma catalytic oxidation of toluene using Au/γ-Al2O3 nanocatalyst. Chem. Eng. J. 427, 130983. https://doi.org/10.1016/j.cej.2021.130983

Xiao, X., Meng, X., Le, D.D., Takarada, T., 2011. Two-stage steam gasification of waste biomass in fluidized bed at low temperature: Parametric investigations and performance
66